\documentclass[10pt,twocolumn,letterpaper]{article}

\usepackage[pagenumbers]{cvpr} 

\usepackage{amsfonts}
\usepackage{amsmath}
\usepackage{amsthm}
\usepackage{booktabs}

\usepackage{bbding}
\usepackage{pifont}
\usepackage[table]{xcolor}

\usepackage{booktabs} 
\usepackage{multirow} 
\usepackage{array} 
\usepackage{makecell} 
\usepackage{algorithm}
\usepackage{algorithmic}

\definecolor{cvprblue}{rgb}{0.21,0.49,0.74}
\usepackage[pagebackref,breaklinks,colorlinks,allcolors=cvprblue]{hyperref}

\def\paperID{15777} 
\def\confName{CVPR}
\def\confYear{2026}

\title{DynamicGTR: Leveraging Graph Topology Representation Preferences \\
to Boost VLM Capabilities on Graph QAs}

\author{Yanbin Wei$^{1,2}$ \quad
Jiangyue Yan$^3$ \quad
Chun Kang$^4$ \quad
Yang Chen$^1$ \quad
Hua Liu$^1$ \quad
James Kwok$^2$ \quad
Yu Zhang$^1$\thanks{Corresponding Author}\\[0.4em]
$^1$Southern University of Science and Technology\quad
$^2$Hong Kong University of Science and Technology\\[0.4em]
$^3$Harbin Institute of Technology (Shenzhen) \quad
$^4$Beihang University\\[0.4em]
{\tt\small yanbin.ust@gmail.com; yu.zhang.ust@gmail.com}
}
\begin{document}
\maketitle
\begin{abstract}
Vision-Language Models (VLMs) have emerged as versatile solutions for zero-shot question answering (QA) across various domains. However, enabling VLMs to effectively comprehend structured graphs and perform accurate, efficient QA remains challenging. Existing approaches typically rely on one single graph topology representation (GTR), such as fixed-style visual images or unified text descriptions. This ``one-size-fits-all'' strategy often neglects model-specific and task-specific preferences, resulting in inaccurate or over-lengthy responses to graph-related queries.
To address this,  we propose the \mbox{DynamicGTR} framework, which dynamically selects the optimal GTR for each query during inference, thereby enhancing the zero-shot graph QA capabilities of VLMs with a customizable accuracy and brevity trade-off. Extensive experiments show that DynamicGTR not only improves VLM-based graph algorithm QA performance but also successfully transfers the experience trained from synthetic graph algorithm tasks to real-world applications like link prediction and node classification, without any additional training. Additionally, DynamicGTR demonstrates strong transferability across tasks, domains, and models, suggesting its potential as a flexible solution for broad graph scenarios.
\end{abstract}    
\section{Introduction}
\label{sec:intro}

\begin{figure}[t]
\centering
\includegraphics[width=0.4\textwidth]{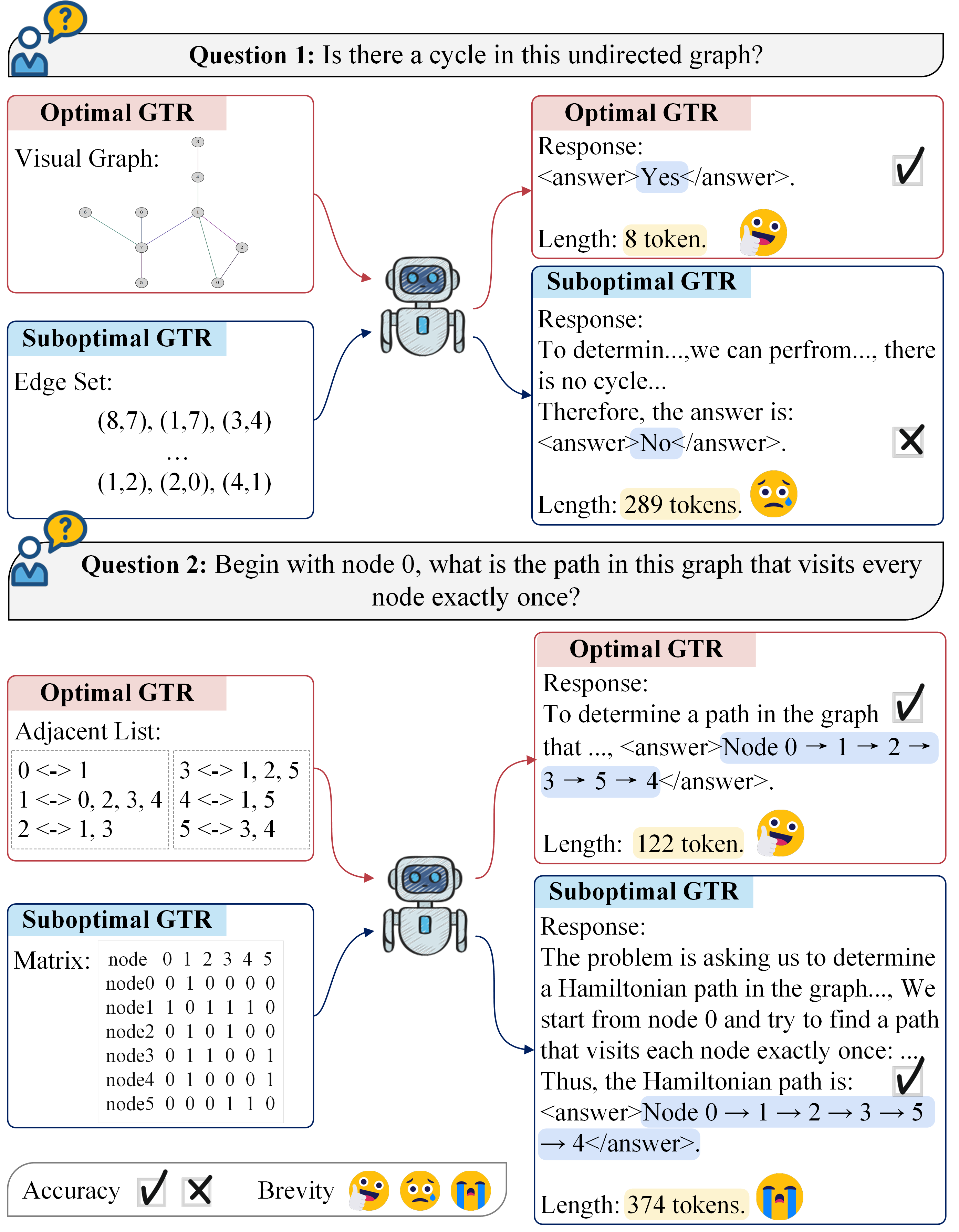}
\caption{Illustrating the impacts of diverse GTRs on the same question: suboptimal GTRs may lead to incorrect answers or unnecessarily lengthy responses compared to a preferred GTR.}
    \label{fig: intro}
    \vspace{-15pt}
\end{figure}

Vision-Language Models (VLMs) have recently demonstrated the ability to perform zero-shot question-answering on structured graphs, answering graph-related questions using only their intrinsic knowledge without task-specific fine-tuning \cite{wei2024gita, li2024visiongraph}. This capability highlights their versatility and potential to handle complex structural data, making them valuable for applications in network analysis \cite{social1, wasserman1994social} and knowledge discovery \cite{wang2014knowledge, huang2021knowledge}.

Existing works have explored various graph topology representations (GTRs) to present the graph topology to VLMs for understanding. Textual GTRs, as demonstrated in \citep{instructGLM, NLGraph, chen2024graphwiz, graphtoken, LLM4Graph}, encode graph structures into textual descriptions using diverse prompt templates. Recently, visual GTRs have been introduced for VLMs \citep{li2024visiongraph, wei2024gita}, depicting graph topologies as stylized images. 

Despite these advancements, existing methods commonly assume the use of a single type of GTR 
(such as unified prompt-based textual descriptions or fixed-style visualizations)
throughout the QA process.
This ``one-size-fits-all'' approach overlooks critical factors, including model-specific cognitive biases and task-specific representational preferences, resulting in suboptimal QA performance for specific tasks (see experiments in Table \ref{tab: ablation}). 
Figure \ref{fig: intro} demonstrate the QAs for the same question with diverse GTRs: A question to identify the cycle in graphs prefers the visual GTR with quick, intuitive pattern recognition over representing the graph with a textual edge set. The suboptimal adjacent matrix GTR leads to prolonged lengthy responses for a path-finding problem. As can be seen, compared to the preferred GTR, suboptimal GTRs can impede the model's comprehension of graph topology, resulting in a wrong answer or prolonged response. This raises intriguing research questions: \textit{Can these preferences for GTRs be leveraged to make graph QA accurate and inexpensive? How and to what extent can they enhance the graph QA capabilities of VLMs?}

In this paper, we present a systematic solution to enhance VLM-based zero-shot graph QA. We begin by categorizing and analyzing the characteristics of existing graph topology representations (GTRs). Next, we design and construct a GTR pool, \(\mathcal{R}_{ZS}\), containing versatile GTRs specifically tailored for zero-shot graph QA, guided by dedicated design principles. We then propose the \textbf{DynamicGTR} framework, which utilizes the GTRs within \(\mathcal{R}_{ZS}\) and consists of three key components:
(1) defining the \textbf{Graph Response Efficiency (GRE)} metric, which balances accuracy and computational cost; (2) leveraging a probe dataset to identify GTRs within \(\mathcal{R}_{ZS}\) that have optimal GRE for each question, thereby constructing a model-specific \textbf{GTR Preference} dataset; (3) training a \textbf{GTR Router} on the GTR Preference dataset to dynamically select appropriate GTRs from \(\mathcal{R}_{ZS}\) for each question during inference. Note that since the GTRs act at the input stage of VLMs, DynamicGTR does not require access to any information about the model architecture or parameters, and thus can be applied to the state-of-the-art closed-source VLMs.

Extensive experiments show that DynamicGTR significantly improves VLM-based graph QA capabilities across seven representative in-domain graph algorithmic tasks. Besides, it can successfully transfer the experience from synthetic graph algorithm tasks to real-world applications like link prediction and node classification, without any additional training. Additionally, the router demonstrates strong transferability across different VLMs and exhibits usability for large graphs, suggesting its strong potential as a flexible solution for broad graph scenarios.

In summary, our key contributions are as follows:
\begin{itemize}
    \item A systematic investigation of existing fixed GTRs, highlighting their characteristics and limitations.
    \item Introduction of DynamicGTR, which adaptively assigns visual or language-based GTRs based on query-specific requirements and user preferences for accuracy and brevity.
    \item The side product GTRP dataset serves as a valuable reference to reveal the mapping from task types to their preferred GTRs.
    \item Empirical results show that DynamicGTR exhibits significant effectiveness on both synthetic graph algorithm QAs and real-world applications, as well as strong generalizability across tasks and models.
\end{itemize}
\section{Related Work}
\label{sec: related}

\paragraph{VLM-based Graph QA.} Graph QA presents unique challenges for VLMs. These challenges stem from the need for structural awareness of graph topology and the ability to select and emulate appropriate algorithms for multi-step reasoning. Existing research in this domain can be broadly categorized into two main approaches:
(1) Toolkit-Enhanced System:
This category encompasses methods such as StructGPT \citep{structgpt}, Graph-Toolformer \citep{zhang2023graph}, and GraphDPR \citep{li2024visiongraph}. These systems utilize predefined external toolkits to assist VLMs in graph QA. Although effective within certain domains, this dependence restricts the rigid question types they can handle, rendering them inadequate for out-of-domain tasks.
(2) Graph-Aware VLMs: Examples in this category include InstructGLM \citep{tang2023graphgpt}, GraphToken \citep{graphtoken}, GraphLLM \citep{chai2023graphllm}, and Gcoder \citep{zhang2024gcoder}. These methods enhance VLMs with graph-awareness by modifying their architecture or fine-tuning their parameters, leveraging the internal knowledge of the VLMs for graph QA. Yet the additional training or architectural change breaks the zero-shot premise and makes deployment infeasible on closed-source models that provide only black-box access.
\section{Zero-shot GTR Pool}
\label{sec: observation}
This section first categorizes and analyzes the existing representative GTRs. It then proposes important principles for designing a Zero-shot GTR pool \(\mathcal{R}_{ZS}\), and constructs an instance of \(\mathcal{R}_{ZS}\) following the proposed principles.

\subsection{Analyzing GTRs in Graph QA} 
\label{sec: analysis}
The GTRs describe the graph topology $G=\{V,E\}$, where $V$ and $E$ denote the set of nodes and edges, respectively. According to the representation types, current GTRs utilized in Graph QA can be categorized into three types: Embedding GTRs, Textual GTRs, and Visual GTRs. In Table \ref{tab: GTR_summary}, we showcase representative GTRs and summarize their characteristics.

\begin{table}[htbp]
    \centering
    \renewcommand{\arraystretch}{1.2} 
    \small 
    \scalebox{0.92}{
    \setlength{\tabcolsep}{1.5pt}
    \begin{tabular}{l|l|c|c|c|c}
        \hline
        \textbf{GTR} & \textbf{Type} & \textbf{Works} & \textbf{Charac.} & \textbf{Enc.} & \textbf{Train} \\
        \hline
        Embeddings & Embed & 
        \begin{tabular}[c]{@{}c@{}}GraphLLM\citep{chai2023graphllm}, \\ GraphGPT\citep{tang2023graphgpt}, \\ GraphToken\citep{graphtoken}, \\ LLaGA\citep{chen2024llaga}\end{tabular} & 
        \begin{tabular}[c]{@{}c@{}}Partial, \\ Informative\end{tabular} & 
        Yes & Yes \\ 
        \hline
        Visual Graph & Visual & 
        \begin{tabular}[c]{@{}c@{}}GITA\citep{wei2024gita}, \\ VisionGraph\citep{li2024visiongraph}\end{tabular} & 
        \begin{tabular}[c]{@{}c@{}}Full, \\ Intuitive, \\ Explicit\end{tabular} & 
        No & No \\
        \hline
        Edge Set & Textual & 
        \begin{tabular}[c]{@{}c@{}}GraphWiz\citep{chen2024graphwiz}, \\ NLGraph\citep{NLGraph}, \\ GPT4Graph\citep{guo2023gpt4graph}, \\
        GraphArena\citep{tang2025grapharena},
        \\
        GCoder\citep{zhang2024gcoder}\end{tabular} & 
        \begin{tabular}[c]{@{}c@{}}Full, \\ Sequential, \\ Implicit\end{tabular} & 
        No & No \\
        \hline
        Adjacent List & Textual & 
        \begin{tabular}[c]{@{}c@{}}InstructGLM\citep{instructGLM}, \\ GraphText\citep{zhao2023graphtext}\end{tabular} & 
        \begin{tabular}[c]{@{}c@{}}Full, \\ Sequential, \\ Implicit\end{tabular} & 
        No & No \\
        \hline
        Adjacent Matrix
        & Textual & 
        \begin{tabular}[c]{@{}c@{}}GraphDPR\citep{li2024visiongraph}\end{tabular} & 
        \begin{tabular}[c]{@{}c@{}}Full, \\ Redundant, \\ Implicit\end{tabular} & 
        No & No \\
        \hline
    \end{tabular}
    }
   \caption{Summary of existing GTRs. `Charac.' indicates the characteristics; `Enc.' and `Train' denote whether the GTR is generated by an external encoder and whether extra training is required, respectively. `Embed' is short for `Embedding'.}
    \label{tab: GTR_summary}
    \vspace{-15pt}
\end{table}

As the table indicates, Embedding GTRs are generated by external encoders and typically require alignment training with the embedding space of VLMs \citep{chai2023graphllm, tang2023graphgpt, chen2024llaga}. While the inevitable compression in embeddings leads to the loss of some topological information \citep{embeddingsurvey}, it also allows the learned embedding GTRs to focus on the most informative patterns.

In contrast, both Visual GTRs and Textual GTRs convey the complete topology of a graph, enabling the graph to be uniquely reconstructed from these representations. However, textual GTRs, whether in the form of an Edge Set or an Adjacency List, present complex structures in a flattened, sequential context. This makes the topological information more implicit compared to the intuitive presentation offered by visual graphs \citep{wei2024gita}. Moreover, the Adjacency Matrix often introduces redundant zero elements for non-existent edges, especially in the case of sparse graphs, making it rarely the primary choice for graph topology representation in existing works. However, the Adjacency Matrix can still be useful in specific scenarios where the explicit representation of connections is advantageous.

\subsection{Constructing the ${R}_{ZS}$}
Given the variety of available GTRs, we carefully select the most suitable ones to form a dedicated set, denoted as \(\mathcal{R}_{\text{ZS}}\), explicitly tailored to zero-shot graph QA scenarios. The selection follows three key principles:
(1) \textit{Model-Agnostic}: The generation of GTRs must be decoupled from the VLM parameters, ensuring compatibility with closed-source VLMs. This is crucial as these models often offer superior performance but do not disclose model details. \textit{Consequently, embedding-based GTRs are excluded from \(\mathcal{R}_{\text{ZS}}\) due to their reliance on embedding spaces, which are inaccessible given most leading models are closed-source}. (2) \textit{Variety}: The GTRs in \(\mathcal{R}_{ZS}\) should exhibit diversity to effectively address a wide range of question types. This diversity helps accommodate different types of graphs and QA tasks, utilizing formats like text and images that naturally align with VLM input formats. (3) \textit{Effectiveness}: Each GTR should possess strong individual capabilities, contributing significantly to the overall QA process. This ensures that each component of the set is valuable and enhances the QA performance.

Following these refined principles, we construct an instance of the zero-shot GTR pool $\mathcal{R}_{ZS} = \{V_{dot}, V_{neato}, V_{circo}, V_{fdp}, V_{sfdp}, T_{set}, T_{list}, T_{mat}\}$, consisting of the following GTRs:

\begin{enumerate}
    \item \textbf{Visual GTRs (5 types):} The GTRs $V_{dot}$, $V_{neato}$, $V_{circo}$, $V_{fdp}$, and $V_{sfdp}$ are generated based on methods from \citep{wei2024gita} using different layout algorithms provided by Graphviz \citep{graphviz}. Specifically, $V_{\text{dot}}$ arranges nodes in tree-like hierarchical layers; $V_{\text{neato}}$ uses the spring model \citep{fruchterman1991graph} to minimize edge crossings on canvas; $V_{\text{circo}}$ positions nodes in a circular pattern; $V_{\text{fdp}}$ offers a fast force-directed layout with optimized computational overhead; and $V_{\text{sfdp}}$ provides a scalable force-directed layout for efficiently handling large graphs.
    
    \item \textbf{Textual GTRs (3 types):} These GTRs $T_{set}$, $T_{list}$, and $T_{mat}$ present the graph topology structure via Edge set, Adjacent List, and Adjacent Matrix, respectively. The prompt templates of $T_{set}$ are retrieved from \citep{NLGraph}, and the others are designed by us.
\end{enumerate}
 We provide the GTRs generation details in Appendix A. 

\begin{figure}[t]
\centering
\includegraphics[width=0.45\textwidth]{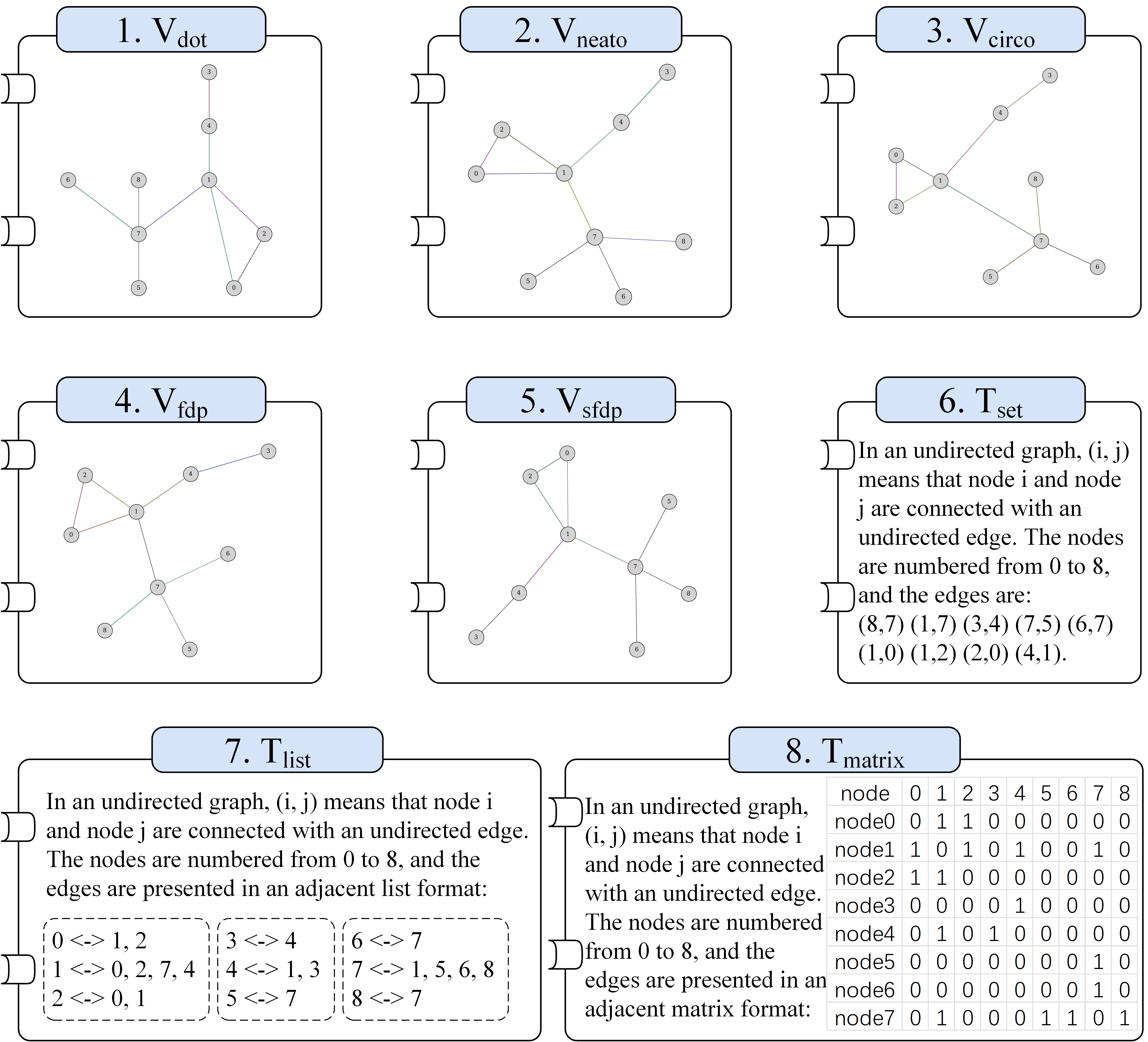}
\caption{An Illustration of eight candidate GTRs in $\mathcal{R}_{ZS}$.}
    \label{fig: examples}
    \vspace{-10pt}
\end{figure}

 Figure \ref{fig: examples} illustrates the examples of GTRs in $\mathcal{R}_{ZS}$, all of which are \textit{Model-agnostic} generated before input to VLMs, ensuring their usability with closed-source models. Visual GTRs enable rapid, intuitive perception of topology, while textual GTRs offer slower, analytical understanding, mirroring dual-system cognitive frameworks \citep{wason1974dual, daniel2017thinking}. Prior works \citep{wei2024rendering, tang2025grapharena} further show that layouts and prompt templates substantially influence graph QA performance, supporting our emphasis on \textit{Variety}. Section \ref{exp: ablation} empirically validates the \textit{effectiveness} of each GTR across tasks.

\begin{figure*}[htbp]
  \includegraphics[width=\textwidth]{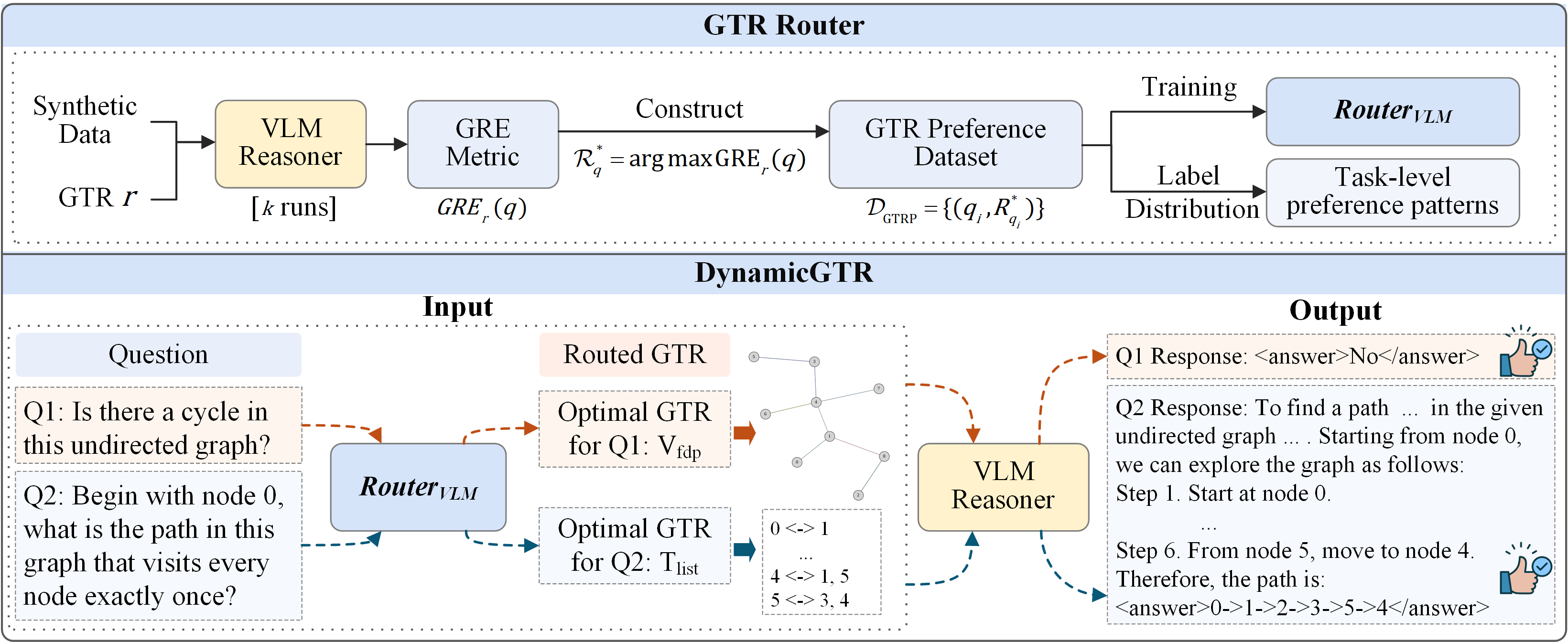}
  \caption{Overview of the DynamicGTR framework, where the GTR Router guides the VLM Reasoner to use the most appropriate GTR based on the question.}
  \label{fig:MD_framework}
  \vspace{-10pt}
\end{figure*}
\section{DynamicGTR}
\label{sec: method}
In this section, we present the DynamicGTR framework, which enhances the zero-shot graph QA of VLMs by dynamic GTR routing, on both accuracy and brevity.

\subsection{Problem Formulation}

VLM-based zero-shot graph QA leverages pretrained VLMs to tackle a wide range of graph problems, without access to or modification of the VLM specifics. This approach facilitates effective generalization across various tasks and unseen graph data, while fully preserving the inherent capabilities of VLMs across the other universal domains.

\subsection{Framework Overview}
\begin{table*}[h!tbp]
    \centering
    \setlength{\tabcolsep}{3pt}
    \renewcommand{\arraystretch}{1.3} 
    \small 
    \begin{tabular}{llccccccc}
        \hline
        Model & GTRs & Conn & Cyc & TS & SP & MF & BGM & HP \\
        \hline
        \multirow{3}{*}{\makecell[c]{GPT-4o}} & 1st 
        &\cellcolor{gray!20}$V_{fdp}$ (92.3\%) 
        &\cellcolor{gray!20} $V_{sfdp}$ (85.1\%)
        & $T_{set}$ (58.5\%)
        & $T_{set}$ (19.5\%)
        & $T_{list}$ (21.7\%)
        &\cellcolor{gray!20} $V_{dot}$ (34.2\%)
        &$T_{list}$ (30.4\%)\\
        \cline{2-9}
        & 2nd & 
        \cellcolor{gray!20}$V_{neato}$ (92.1\%) 
        & \cellcolor{gray!20}$V_{fdp}$ (12.9\%)
        & $T_{list}$ (36.2\%)
        &\cellcolor{gray!20} $V_{neato}$ (18.7\%)
        & $T_{set}$ (20.8\%)
        &\cellcolor{gray!20} $V_{circo}$ (20.3\%)
        & $T_{set}$ (20.3\%)\\ 
        \cline{2-9}
        & 3rd 
        &\cellcolor{gray!20} $V_{sfdp}$ (91.7\%)
        &\cellcolor{gray!20} $V_{dot}$ (12.2\%)
        &\cellcolor{gray!20} $V_{dot}$ (23.1\%)& $T_{list}$ (17.1\%)& $T_{mat}$ (16.7\%)
        &\cellcolor{gray!20} $V_{neato}$ (19.8\%)
        &\cellcolor{gray!20} $V_{circo}$ (18.8\%)\\
        \hline

        \multirow{3}{*}{\makecell[c]{Gemini \\ 2.5 Pro}} & 1st 
        &\cellcolor{gray!20} $V_{neato}$ (88.8\%)
        &\cellcolor{gray!20}
        $V_{neato}$ (97.2\%)
        &
        $T_{set}$ (41.0\%)
        &
        $T_{list}$ (48.6\%)
        &
        $T_{mat}$ (40.0\%)
        &\cellcolor{gray!20}
        $V_{fdp}$ (24.4\%)
        &
        $T_{list}$ (42.9\%)\\
        \cline{2-9}
        & 2nd &\cellcolor{gray!20}
        $V_{fdp}$ (88.2\%)
        & \cellcolor{gray!20} $V_{sfdp}$ (93.0\%)
        & $T_{list}$ (30.3\%)
        & $T_{mat}$ (37.6\%)
        & $T_{list}$ (36.0\%)
        & \cellcolor{gray!20} $V_{sfdp}$ (14.5\%)
        & $T_{set}$ (31.7\%)
        \\
        \cline{2-9}
        & 3rd 
        &\cellcolor{gray!20} $V_{sfdp}$ (88.2\%)
        & \cellcolor{gray!20} $V_{dot}$ (71.8\%)
        & $T_{mat}$ (15.4\%)
        & $T_{set}$ (26.6\%)
        & $T_{set}$ (14.0\%)
        & \cellcolor{gray!20} $V_{neato}$ (14.5\%)
        & $T_{mat}$ (30.2\%)
        \\
        \hline
    \end{tabular}
    \caption{Task-preferred Top-3 GTRs with frequency in GTRP dataset (GPT-4o and Gemini-2.5 Pro). `Conn', `Cyc', `TS', `SP', `MF', `BGM', and `HP' denote connectivity, cycle detection, topological sort, shortest path, maximum flow, bipartite graph matching, and Hamilton path. By differing visual or textual GTRs with colors, \textbf{the special preference patterns of tasks are explicitly exposed}.}
    \vspace{-5pt}
    \label{tab: preference}
    \vspace{-10pt}
\end{table*}
The DynamicGTR framework, shown in Figure \ref{fig:MD_framework}, establishes a graph QA system with question-specific dynamic GTRs. It consists of two main components: (1) a VLM Reasoner for zero-shot graph QA, and (2) a GTR Router for adaptive GTR selection.

DynamicGTR defines a Graph Response Efficiency (GRE) score to evaluate the trade-off between accuracy and computational cost of GTRs concerning the model and problem context. Using fixed probe data, DynamicGTR identifies the mapping from questions to their optimal GTRs with the highest GRE, creating a GTR Preference dataset to train the GTR Router $Router_{VLM}$.

Then, for any input question $q$, the GTR Router dynamically assigns the most suitable GTR $r_q \in \mathcal{R}_{ZS}$ and the VLM Reasoner uses $r_q$ as its GTR input to perform zero-shot inference, producing the answer $ans_q^{r_q}$. 

\subsection{Graph Response Efficiency (GRE)}

In this section, we introduce the \textit{Graph Response Efficiency (GRE)} score, designed to balance the trade-off between accuracy and computational cost across various GTRs.

For a given question $q$ and a GTR $r \in \mathcal{R}_{ZS}$ utilized by the VLM Reasoner, the \textit{Graph Response Efficiency} is defined as a linear combination of an accuracy objective $\text{Acc}_r(q)$ and an efficiency objective $\text{Eff}_r(q)$:

\begin{equation}
\label{definition: gre}
    GRE_r(q) = \text{Acc}_r(q) + \alpha \times \text{Eff}_r(q),
\end{equation}

The accuracy objective, $\text{Acc}_r(q) = \log(1 + 100 \times \text{correctness}_r(q))$, where $\text{correctness}_r(q) \in \{0,1\}$, represents the log-transformed answer correctness. The efficiency objective, $\text{Eff}_r(q) = -\log(\text{tok}_r(q))$, is the negative logarithm of the average token consumption of the response from the VLM Reasoner using GTR $r$.

\noindent\textit{Design Principle of the log transform.}  
The log transformation is employed in both accuracy and efficiency objectives for several reasons. First, it compresses the scale of metric variation, reducing the impact of outliers and making the score more robust to extreme values or anomalously large token consumption. Second, the logarithmic form accentuates significant improvements in correctness and efficiency at the lower end of the scale, while marginal gains at higher values have reduced influence. For correctness, $\log(1 + 100 \times \text{correctness}_r(q))$ ensures that a correct answer ($=1$) receives a substantial boost compared to an incorrect one ($=0$), reinforcing the importance of factual correctness, especially in single-shot settings. For efficiency, $-\log(\text{tok}_r(q))$ penalizes longer responses exponentially, thereby incentivizing more concise outputs from the VLM Reasoner.

\noindent\textit{Design Principle of Hyperparameter $\alpha$.} The hyperparameter $\alpha$ provides users with the ability to quantitatively adjust the balance between accuracy and brevity (efficiency). For example, a user or setting that prioritizes accuracy over response brevity can opt for a small $\alpha$, such as 0 in extreme cases, effectively removing the impact of brevity on the GRE score. Conversely, users who prioritize efficiency can select a larger $\alpha$. This customizable design enhances the framework's flexibility to accommodate diverse user or scenario requirements.


\subsection{GTR Preference Dataset}
\label{sec: GTRp}
We construct the \textit{GTR Preference Dataset} to systematically map each graph QA question to its empirically optimal topology representation forms (GTRs), as measured by the GRE score. Specifically, we generate 7K graph QA pairs across seven representative graph algorithms (as detailed in Section~\ref{exp:setup}), with each graph topology $G=(V, E)$ sampled from an Erdős–Rényi model \citep{er_model} (varying $\text{node count }N \in [3,30], \text{edge probability} \in [0.1,0.7]$\footnote{Insights from small-scale graph probing successfully transfer to large-scale applications, as demonstrated in Section \ref{ood analysis}.})
Response correctness is verified using strictly-guaranteed algorithmic solutions.
For each question $q$, we define its preferred GTRs $\mathcal{R}^*_q$ as:
\begin{equation}
\mathcal{R}^*_q = \arg\max_{f \in \mathcal{R}_{ZS}} \text{GRE}_f(q),
\end{equation}
where $\text{GRE}_f(q)$ is averaged over $k$ trials to ensure robustness of the preference estimation. Pairing each question $q$ with its corresponding set of preferred GTRs $\mathcal{R}^*_q$, we form the GTR Preference dataset, denoted as $\mathcal{D}_{\text{GTRP}} = \{(q_i, \mathcal{R}_{q_i}^*)\}$.
More details of the GTRP dataset construction method are in Appendix B. 

The GTR Preference dataset is a crucial resource for examining the task-level preferences of GTRs. Analyzing the label frequency statistics for specific tasks can reveal the task-level GTR preferences. Table \ref{tab: preference} provides the top-3 GTR choices in various tasks of the GTRP dataset, with their frequency to be selected in $\mathcal{R}^*_q$ listed in parentheses.

We can conclude findings from Table \ref{tab: preference} by categorizing the 7 tasks into three types: \textbf{(1) Perceptual-Intensive Tasks}: Visual GTRs dominate tasks such as Connectivity, Cycle Detection, and Bipartite Graph Matching. These tasks typically require fast and intuitive topology-awareness, which visual representations are well-suited for.
\textbf{(2) Edge-Weighted Tasks}: Tasks involving edge weights, like Shortest Path and Maximum Flow, prefer textual GTRs. This preference indicates that textual representations are more analytical and suitable for computation-heavy processes. \textbf{(3) Ordered Decomposition Tasks}: Hamilton Path and Topological Sorting require ordered decomposition of the graph, and these tasks also prefer textual GTRs. This suggests that textual representations facilitate structured and sequential processing. We include the complete ranking of GTR labels with frequency statistics and more analysis in Appendix C.

\subsection{GTR Router}
\label{mind_router}
The GTR Router serves as the core decision-making module, dynamically selecting an appropriate GTR \(r_q \in \mathcal{R}_{ZS}\) for each question $q$. 
As defined in Section \ref{definition: gre}, the GRE is a scalarized objective that integrates accuracy (\(\text{Acc}_r\)) and efficiency (\(\text{Eff}_r\)) for dual-objective optimization.
To obtain the GTR Router that approximates the GRE-optimal routing $R^*$, we treat it as a classification model $\text{R}_\phi(q): \mathcal{Q}\mapsto\mathcal{R}_{ZS}$ and train it on the GTRP dataset $\mathcal{D}_{\text{GTRP}} = \{(q_i, R^*_{q_i})\}$. For each GTR $r \in \mathcal{R}_{ZS}$, we define $y_r$ as an indicator of whether $r$ is in the true label set $R^*_q$:
\( 
y_r = \mathbb{I}[r \in R^*_q].
\)
The loss function is then defined as:
\begin{align*}
\mathcal{L}(\phi) = -\mathbb{E}_{(q,R^*_q)\sim\mathcal{D}_{\text{GTRP}}} \Bigg[ & \sum_{r \in \mathcal{R}_{ZS}} y_r \log p_\phi(y_r|q) \\
& + (1 - y_r) \log (1 - p_\phi(y_r|q)) \Bigg],
\end{align*}
where $p_\phi(y_r|q)$ represents the probability that GTR $r$ is present in the true label set $R^*_q$. We adopt DeBERTaV3-base \citep{he2021debertav3} as the router in our experiments. Notably, the router training is lightweight, which only costs $\sim$2.96 hours on one NVIDIA A100 GPU.

\section{Experiment}
In this section, we empirically evaluate the proposed DynamicGTR framework.

\begin{table*}[h!tbp]
    \centering
    \setlength{\tabcolsep}{1.5pt}
    \renewcommand{\arraystretch}{0.9} 
    \small 
    \begin{tabular}{cccccccccccccccc}
        \hline
        & \multicolumn{2}{c}{\raisebox{-0.9ex}[0pt]{Conn}} 
        & \multicolumn{2}{c}{\raisebox{-0.9ex}[0pt]{Cyc}} 
        & \multicolumn{2}{c}{\raisebox{-0.9ex}[0pt]{TS}} 
        & \multicolumn{2}{c}{\raisebox{-0.9ex}[0pt]{SP}} 
        & \multicolumn{2}{c}{\raisebox{-0.9ex}[0pt]{MF}} 
        & \multicolumn{2}{c}{\raisebox{-0.9ex}[0pt]{BGM}} 
        & \multicolumn{2}{c}{\raisebox{-0.9ex}[0pt]{HP}} \\
        Method
        &\multicolumn{2}{c}{\raisebox{0.5ex}{\rule{2.05cm}{0.4pt}}}
        &\multicolumn{2}{c}{\raisebox{0.5ex}{\rule{2.05cm}{0.4pt}}}
        &\multicolumn{2}{c}{\raisebox{0.5ex}{\rule{2.05cm}{0.4pt}}}
        &\multicolumn{2}{c}{\raisebox{0.5ex}{\rule{2.05cm}{0.4pt}}}
        &\multicolumn{2}{c}{\raisebox{0.5ex}{\rule{2.05cm}{0.4pt}}}
        &\multicolumn{2}{c}{\raisebox{0.5ex}{\rule{2.05cm}{0.4pt}}}
        &\multicolumn{2}{c}{\raisebox{0.5ex}{\rule{2.05cm}{0.4pt}}}
        \\
         & Acc & Tok & Acc & Tok & Acc & Tok & Acc & Tok & Acc & Tok & Acc & Tok & Acc & Tok \\
        \midrule
        \multicolumn{15}{c}{\textit{GPT-4o}} \\
        \midrule
        CoT 
        & 92.5 & \underline{273.3}
        & 52.7 & 480.6
        & 36.6 & 224.2 
        & 54.6 & 566.0 
        & 25.3 & 362.9
        & 69.5 & 370.1
        & 50.0 & \underline{124.9}
        \\
        NLGraph 
        & 92.9 & 296.6
        & 60.2 & 337.9 
        & 36.2 & \underline{202.6}
        & 59.0 & 533.7
        & 25.6 & \textbf{335.1} 
        & 62.2 & \underline{356.3}
        & 57.1 & 176.5
        \\
        GraphDPR
        & \underline{94.3} & 412.2
        & \underline{68.7} & 626.4
        & 40.2 & 411.0
        & \underline{59.1}&\underline{496.3}
        & \underline{31.2}&502.7
        & 76.5&582.1
        & \underline{62.7}&475.8
        \\
        GITA
        & 93.4&285.3
        & 64.7&\underline{325.7}
        & \textbf{40.9}&256.1
        & 56.8&\textbf{482.1}
        & 27.4&\underline{359.9}
        & \underline{82.5}&392.4
        & 54.9&188.8
        \\
        \rowcolor{gray!20} 
        DynamicGTR
        & \textbf{96.1} & \textbf{38.8} 
        & \textbf{89.3} & \textbf{75.9}
        & \underline{40.8} & \textbf{176.1}
        & \textbf{68.4}& 499.1
        & \textbf{36.6}&385.2
        & \textbf{92.0}&\textbf{233.6}
        & \textbf{63.3}&\textbf{75.5}
        \\
        \midrule
        \multicolumn{15}{c}{\textit{Gemini-2.5 Pro}} \\
        \midrule
        CoT
        &97.2 & \underline{218.7}
        &98.6 & 716.6
        &84.1 & \underline{1395.9}
        &93.6 & 810.8
        &91.7 &1154.9
        &97.1 & 1076.8
        &96.8 & \underline{678.1}
        \\
        NLGraph 
        & 97.3 & 220.0
        & 98.6 & 629.9
        & 84.5 & 1437.8
        & 94.0 & 847.5
        & 92.2 & \underline{1062.9}
        & 97.1 & \underline{1026.4}
        & 97.1 & 742.9
        \\
        GraphDPR
        & \underline{98.5} & 396.6
        & 98.9 & 907.7
        & \underline{86.2} & 1651.4
        & \underline{95.5} & 849.5
        & \underline{94.8} & 1319.6
        & 97.8 & 1187.2
        & \underline{97.5} & 992.9
        \\
        GITA
        & 98.2&238.5
        & \underline{99.2}&\underline{478.9}
        & 85.5 & 1398.0
        & 94.3 & \underline{808.2}
        & 94.2 & 1155.1
        & \underline{99.1} & 1078.2
        & 97.3 & 679.8
        \\
        \rowcolor{gray!20} 
        DynamicGTR
        &\textbf{100} & \textbf{12.9}
        &\textbf{99.3} & \textbf{16.7}
        &\textbf{87.8} & \textbf{1191.2}
        &\textbf{96.3} & \textbf{798.6}
        &\textbf{100} & \textbf{1004.8}
        &\textbf{100} & \textbf{776.0}
        &\textbf{100} & \textbf{254.6}
        \\
        \hline
    \end{tabular}
    \caption{Zero-shot capabilities on in-domain graph algorithmic tasks. Acc and Tok refer to task-average accuracy (\%) and token consumption, respectively.}
    \label{tab: main1}
    \vspace{-10pt}
\end{table*}

\subsection{Experimental Setup}
\label{exp:setup}

\paragraph{Tasks and Datasets.}
\label{exp:datasets}

We assess the zero-shot graph QA capabilities of DynamicGTR in two distinct settings:

\begin{itemize}
    \item \textbf{In-Domain Algorithm QAs}: We focus on comprehensive graph algorithmic tasks such as Connectivity, Cycle, Topological Sorting, Shortest Path, Maximum Flow, Bipartite Graph Matching, and Hamilton Path (abbreviated as 'Conn', 'Cyc', 'TS', 'SP', 'MF', 'BGM', and 'HP', respectively) \citep{sedgewick2001algorithms, kahn1962topological, dijkstra2022note, ford1956maximal, karp1990optimal, gould2003advances}. The data for these tasks is sourced from the GVLQA-BASE benchmark \cite{wei2024gita}. These tasks encompass a broad spectrum of essential skills for understanding and reasoning about graph topology and align with GTR router training.

  \item \textbf{Out-of-Domain Real-World Applications}: We examine the framework's out-of-domain generalizability and real-world application value by focusing on link prediction (LP) and node classification (NC), two fundamental tasks in the graph domain. For LP, we utilize the collaboration networks ca-GrQC and ca-HepTh \citep{cadataset}, along with 15K queries from the large-scale protein-protein interaction dataset ogbl-ppa \cite{ogb}. For NC, we employ the politician blog network PolBlog \citep{adamic2005political}, the citation network Cora \citep{yang2016revisiting}, and 15K queries of the extensive e-commerce dataset ogbn-product \cite{ogb}. To enhance clarity in text and visual GTRs, and adhering to the locality principle\footnote{Locality Principle: Correlated information is contained within nearby neighbors \citep{wu2020comprehensive, bronstein2017geometric}.}, we depict graph topology using n-hop node-surrounding sampled subgraphs instead of entire graph inputs, following established subgraph-based research practices \citep{kipf2016semi, hamilton2017inductive, xu2018powerful, velickovic2017graph, chen2022structure}. Unlike the algorithm QAs where the GTR router was trained, these real-world applications feature entirely novel domains with complex semantics and much larger graphs. More details about task specifics and data statistics are in Appendix D.
\end{itemize}

\begin{table*}[tbp]
    \centering
    \setlength{\tabcolsep}{5.5pt}
    \renewcommand{\arraystretch}{1.0} 
\small 
\begin{tabular}{lcccccccccccc}
    \hline
    & \multicolumn{2}{c}{\raisebox{-0.9ex}[0pt]{ca-GrQC}} 
    & \multicolumn{2}{c}{\raisebox{-0.9ex}[0pt]{ca-HepTh}} 
    & \multicolumn{2}{c}{\raisebox{-0.9ex}[0pt]{ogbl-ppa}} 
    & \multicolumn{2}{c}{\raisebox{-0.9ex}[0pt]{PolBlog}} 
    & \multicolumn{2}{c}{\raisebox{-0.9ex}[0pt]{Cora}} 
    & \multicolumn{2}{c}{\raisebox{-0.9ex}[0pt]{ogbn-product}}  \\
    Method
    &\multicolumn{2}{c}{\raisebox{0.5ex}{\rule{2.05cm}{0.4pt}}}
    &\multicolumn{2}{c}{\raisebox{0.5ex}{\rule{2.05cm}{0.4pt}}}
    &\multicolumn{2}{c}{\raisebox{0.5ex}{\rule{2.05cm}{0.4pt}}}
    &\multicolumn{2}{c}{\raisebox{0.5ex}{\rule{2.05cm}{0.4pt}}}
    &\multicolumn{2}{c}{\raisebox{0.5ex}{\rule{2.05cm}{0.4pt}}}
    & \multicolumn{2}{c}{\raisebox{0.5ex}{\rule{2.05cm}{0.4pt}}}
    \\
     & Acc & Tok & Acc & Tok & Acc & Tok & Acc & Tok & Acc & Tok & Acc & Tok \\
    \midrule
    GPT-4o 
    & 66.2 & \underline{191.2}
    & 77.4 & \underline{229.8}
    & 42.6& \underline{199.4}
    & 59.2& 255.7
    & 52.2& 228.5
    & 33.2& 192.6
    \\ 
     \ +CoT
     & 68.6  & 244.0
     & 77.8 & 268.6
     & 48.2& 238.1
     & 60.3 & 311.2
     & 54.4& 272.2
     & 33.1& 199.9 
    \\
     \ +GITA
     & \underline{73.1} & 203.1
     & \underline{83.1}& 245.9
     & \underline{52.8}& 266.2 
     & \underline{63.6}& \underline{226.6}
     & \underline{56.8}& \underline{184.4}
     & \underline{35.5}& 233.4 
    \\ 
     \rowcolor{gray!20}\ +DynamicGTR
     & \textbf{76.5} & \textbf{143.2}
    & \textbf{86.3} &  \textbf{183.4}
    & \textbf{54.4} & \textbf{206.3}
    & \textbf{71.9}& \textbf{153.0}
    & \textbf{64.5}& \textbf{99.2}
    & \textbf{40.1}& \textbf{156.1}\\
    \midrule
    Gemini-2.5 Pro
    & 71.9 & \underline{278.5}
    & 82.5 & 321.7
    & 57.2& \underline{322.7}
     & 63.7 & 312.6
     & 57.3 & 348.4
     & 40.8 & \underline{208.8}
    \\ 
     \ +CoT
     & 72.1& 316.0
     & 82.9& 345.1
     & 59.4& 338.7
     & 64.3& 332.7
     & 57.5& 388.9
     & 42.5 & 229.9
    \\
     \ +GITA 
     & \underline{74.8}& 301.1
     & \underline{85.8}& \underline{319.3}
     & \underline{62.1} & 365.9
     & \underline{65.7}& \underline{298.6}
     & \underline{59.9}& \underline{306.2}
     & \underline{45.6} & 211.2
    \\ 
     \rowcolor{gray!20}\ +DynamicGTR
     & \textbf{77.9}& \textbf{198.8}
    & \textbf{89.7}& \textbf{251.3}
    & \textbf{64.8}& \textbf{189.2}
    & \textbf{76.4}& \textbf{221.9}
    & \textbf{68.6}& \textbf{145.3}
    & \textbf{51.9} & \textbf{155.0}\\
    \hline
\end{tabular}
    \caption{Zero-shot capacities on out-of-domain real-world applications across six datasets. Acc and Tok refer to task-average accuracy (\%) and token consumption, respectively.}
    \label{tab: main2}
    \vspace{-15pt}
\end{table*}
\begin{table*}[t]
    \centering
    \setlength{\tabcolsep}{1.5pt}
    \renewcommand{\arraystretch}{0.8} 
    \small 
    \begin{tabular}{lcccccccccccccccc}
        \hline
        &  & \multicolumn{2}{c}{\raisebox{-0.6ex}[0pt]{Conn}} 
        & \multicolumn{2}{c}{\raisebox{-0.6ex}[0pt]{Cyc}} 
        & \multicolumn{2}{c}{\raisebox{-0.6ex}[0pt]{TS}} 
        & \multicolumn{2}{c}{\raisebox{-0.6ex}[0pt]{SP}} 
        & \multicolumn{2}{c}{\raisebox{-0.6ex}[0pt]{MF}} 
        & \multicolumn{2}{c}{\raisebox{-0.6ex}[0pt]{BGM}} 
        & \multicolumn{2}{c}{\raisebox{-0.6ex}[0pt]{HP}} 
        & \\
        GTR & Avg.GRE
        &\multicolumn{2}{c}{\raisebox{0.5ex}{\rule{2.05cm}{0.4pt}}}
        &\multicolumn{2}{c}{\raisebox{0.5ex}{\rule{2.05cm}{0.4pt}}}
        &\multicolumn{2}{c}{\raisebox{0.5ex}{\rule{2.05cm}{0.4pt}}}
        &\multicolumn{2}{c}{\raisebox{0.5ex}{\rule{2.05cm}{0.4pt}}}
        &\multicolumn{2}{c}{\raisebox{0.5ex}{\rule{2.05cm}{0.4pt}}}
        &\multicolumn{2}{c}{\raisebox{0.5ex}{\rule{2.05cm}{0.4pt}}}
        &\multicolumn{2}{c}{\raisebox{0.5ex}{\rule{2.05cm}{0.4pt}}}
        \\
               &  & Acc & Tok & Acc & Tok & Acc & Tok & Acc & Tok & Acc & Tok & Acc & Tok & Acc & Tok \\
        \midrule
        $V_{dot}$ & 1.16
        &78.5 & 8.4
        &80.0 & \underline{8.0}
        &12.3 & 346.0
        &14.5&\underline{146.3}
        &7.5&410.2
        &\underline{91.2}&\underline{244.3}
        &13.3&\textbf{40.0}
        \\
        $V_{neato}$ & 1.20
        &95.1&\textbf{7.7}
        &\underline{87.1}&\underline{8.0}
        &3.0&\textbf{30.0}
        &20.0&\textbf{130.1}
        &10.8&387.4
        &87.2&280.0
        &11.1&\textbf{40.0}
        \\
        $V_{circo}$ & 1.13
        &89.7& 8.2
        &70.7&\underline{8.0}
        &3.0&\textbf{30.0}
        &18.2&153.5
        &8.1&421.3
        &88.2&271.6
        &14.4&\textbf{40.0}
        \\
        $V_{fdp}$ & 0.88
        &\underline{96.0}&\underline{8.1}
        &60.5&8.0
        &3.0&\textbf{30.0}
        &17.3&150.5
        &8.1&389.2
        &82.9&295.3
        &4.4&74.0
        \\
        $V_{sfdp}$ & 1.21
        &94.8&\underline{8.1}
        &85.1&\textbf{7.0}
        &7.0&120.0
        &25.5&168.7
        &8.1&386.1
        &84.0&302.9
        &12.2&40.0
        \\
        $T_{set}$ & 1.05
        & 92.5&273.3
        & 52.7&480.6
        & \underline{36.6}&224.2
        & 54.6&566.0
        & \underline{25.3}&\textbf{362.9}
        & 69.5&370.1
        & \underline{50.0}&124.9
        \\
        $T_{list}$ & 1.07
        & 89.0&218.2
        & 53.2&359.7
        & 34.2&206.3
        & \underline{55.5}&518.9
        & 24.2&400.7
        & 65.8&410.8
        & 50.0&107.0
        \\
        $T_{mat}$ & 0.56
        &79.9&233.7
        &52.7&417.1
        &6.8&378.1
        &34.5&591.8
        &17.2&402.9
        &60.4&407.3
        &31.1&160.9
        \\
        \rowcolor{gray!20}  Router & \textbf{1.67}
        & \textbf{96.1}&38.8 
        & \textbf{89.3}&75.9
        & \textbf{41.4}&176.1
        & \textbf{68.4}&499.1
        & \textbf{36.6}&\underline{385.2}
        & \textbf{92.0}&\textbf{233.6}
        & \textbf{61.1}&76.3
        \\
        \hline
    \end{tabular}
    \caption{Comparison of the DynamicGTR framework with GTR Router versus single GTRs on in-domain tasks (GPT-4o). Additional results for more VLMs in Appendix G show similar trends.}
    \label{tab: ablation}
    \vspace{-5pt}
\end{table*}
\paragraph{Baselines.} We evaluate the zero-shot graph QA capabilities of DynamicGTR against the following baselines for in-domain tasks: (1) Vanilla Chain-of-Thought (CoT) \citep{wei2022chain}, which uses step-by-step prompts; 
(2) NLGraph \citep{NLGraph} that utilizes BAG prompting to conceptualize a graph and algorithmic prompting to specify the algorithm;
(3) GraphDPR \citep{li2024visiongraph} employs external tools to generate intermediate descriptions and code with multi-step reasoning;
(4) GITA \citep{wei2024gita} that pairs a visual GTR and a textual GTR to collaboratively solve graph algorithmic problems. For monetary cost, we only evaluate all methods with leading closed-source VLMs GPT-4o \citep{gpt4o} and Gemini-2.5 Pro \citep{gemini2.5}. The proposed method can also work on open-source VLMs, we provide more results on LLaVA-OneVision-7B \citep{li2024llavaonevision} and Qwen3-VL-8B \citep{qwen3} in Appendix E. For each question, we evaluate three times with temperature $\tau=0.7$ and report the task-average accuracy (noted Acc\%) and token cost (noted Tok) across trials. For out-of-domain tasks, we use the base model, CoT, and GITA as baselines\footnote{Methods like NLGraph and GraphDPR are not applicable due to the absence of definitive algorithms for downstream tasks.}. By default, we assume a relatively neutral user preference between accuracy and efficiency with setting the trade-off parameter $\alpha$ in GRE score as $0.5$ \footnote{Discussions for more user trade-off preferences with varying the $\alpha$ are provided in Section \ref{sec:exp_hyper}.}. Implementation details of all methods are in Appendix F.

\subsection{Main Results}
\paragraph{In-Domain Algorithm QAs.}
Table \ref{tab: main1} compares the effectiveness (accuracy) and efficiency (token consumption) of DynamicGTR against baseline methods across seven in-domain graph algorithm QA tasks. As stated in Section \ref{sec: GTRp}, these tasks acquire three types of distinct graph capabilities. We also arrange our findings with analysis according to such categorization: (1) For perceptual-intensive tasks (Conn, Cyc, BGM), DynamicGTR significantly enhances accuracy while reducing token consumption. These tasks demand rapid and intuitive perception, favoring visual GTRs that excel in both accuracy and token efficiency, thus enabling effective and swift reasoning. (2) For edge-weighted tasks (SP, MF), the benefits of DynamicGTR are more reflected in accuracy over token consumption. This is because the fulfillment of these tasks typically relies on analytical computation, where the reasoning processes play an important role and bring a necessary token cost. (3) For ordered decomposition tasks (TS, HP), the framework plays more roles in saving token consumption over increasing accuracy. Specifically, in these tasks, we observe that the less prevalent GTRs selected in rare ($<15\%$) cases contribute a majority ($>88\%$) of token savings, revealing the diverse sample-wise demands within these tasks.  
\paragraph{Out-of-domain Real-world Applications.}
\label{ood analysis}
Table \ref{tab: main2} presents a comprehensive comparison of DynamicGTR and baseline methods on link prediction (LP) and node classification (NC) across six diverse real-world graph datasets. Notably, both LP and NC represent tasks and domains that were entirely unseen during GTR router training. Despite this domain and task shift, DynamicGTR consistently outperforms baselines in both accuracy and token efficiency, even for challenging large-scale graphs such as ogbl-ppa and ogbl-product. Crucially, these results highlight DynamicGTR’s robust transferability: The router trained solely on synthetic algorithmic QAs on small-scale graphs can adaptively select optimal GTRs for previously unseen complex tasks, diverse domains, and much larger graph sizes without additional fine-tuning. This demonstrates that the selected graph algorithms are good gyms for DynamicGTR to capture essential and common GTR preference patterns across scenarios, and the DynamicGTR routing obsesses generalizable application value to promote VLM-based graph applications.

\subsection{Ablation Study}
\label{exp: ablation}
To elucidate the necessity of the proposed GTR Routing, we present metrics for individual GTRs in Table \ref{tab: ablation}, compared with the proposed GTR Router within the DynamicGTR framework. As can be seen, there does not exist any GTR that can dominate the others across tasks. For each task, the top GTRs are highly aligned with the task preferences in the GTRP dataset (Table \ref{tab: preference}). With the GTR Router, DynamicGTR achieves the highest Avg.GRE scores across all tasks beyond all individual GTRs, underscoring that it faithfully improves the overall performance according to the current-specified accuracy-efficiency tradeoff.

\subsection{Hyperparameter Effects Analysis}
\label{sec:exp_hyper}
In this section, we discuss the effects of hyperparameters in the framework. Table \ref{tab: alpha} and \ref{tab: k} demonstrate the performance effects of varying $\alpha$ and $k$, respectively, where the reported values are averaged separately on 1) all in-domain algorithm QAs (I.D.). 2) Link prediction (LP) benchmarks. 3) Node classification (NC) benchmarks. For direct comparison, we also list the best performances of the non-DynamicGTR baselines with respect to each value noted as `Best Baseline' for reference. Due to space limits, we show GPT-4o results here (Gemini-2.5 Pro exhibits the same trends and is included in Appendix H).
\paragraph{Effects on $\alpha$.}
Table \ref{tab: alpha} demonstrates how varying $\alpha$ values in Eq.(\ref{definition: gre}) influences overall performance.
(1) Compared with $\alpha=0.5$, a lower value ($\alpha=0$) makes GRE dominated by accuracy, thereby increasing QA accuracy but at the cost of higher token consumption. Conversely, a higher value ($\alpha=1$) strongly penalizes token usage, resulting in greater efficiency but reduced accuracy. This allows users to effectively control the tradeoff between accuracy and efficiency simply by adjusting $\alpha$, making DynamicGTR flexible for different deployment scenarios.
(2) DynamicGTR outperforms the best baselines in both dimensions across all $\alpha$ values in most cases. As the reflection of their preferences, $\alpha=0$ and $\alpha=1$ always achieve the best accuracy and efficiency, respectively. (3) Furthermore, when a user wants to adjust their preference for accuracy and efficiency, this only requires updating $\alpha$ and recomputing the GTR preferences (i.e., GTRP dataset) based on the previously dumped probed results without incurring any further API cost. In our practices, the lightweight retraining of GTR Router can be completed within 3 GPU-hours on one NVIDIA A100, ensuring rapid, cost-effective adaptation to new user preferences.
\begin{table}[h!tbp]
    \centering
    \vspace{-5pt}
    \small
    \setlength{\tabcolsep}{8pt}
    \renewcommand{\arraystretch}{0.78} 
        \begin{tabular}{lllll}
            \toprule
            \textbf{Metric} 
            & setting 
            & I.D. 
            & LP 
            & NC 
 \\
            \midrule

            & $\alpha=0$ &  \textbf{69.6} & \textbf{73.1} & \textbf{59.3}\\
            
        Acc$\uparrow$ & $\alpha=0.5$ & \underline{69.5} & \underline{72.4} & \underline{58.8} \\
            
            & $\alpha=1$ & 69.3 & 72.0 & 58.6\\
            & Best Baseline & 61.8 & 69.7 & 52.0 \\
            \midrule

            &  $\alpha=0$ &  290.8 & 223.2 & 168.8 \\
        Tok$\downarrow$ & $\alpha=0.5$ & \underline{212.0} & \underline{177.6} & \underline{136.1}\\
            &  $\alpha=1$ & \textbf{209.9} & \textbf{174.1} & \textbf{133.9} \\
            & Best Baseline & 319.8 & 206.8 & 214.8\\


            \bottomrule
        \end{tabular}
        \vspace{-5pt}
        \caption{ The accuracy and token consumption with setting different $\alpha$ values. The base VLM is GPT-4o.}
        \label{tab: alpha}
        \vspace{-20pt}
\end{table}
\paragraph{Effects on 
$k$.} Table \ref{tab: k} shows the impact of the hyperparameter $k$. By repeating inferences across trials, increasing 
$k$ improves the quality of the GTR preference dataset by aligning probing results more closely with the true GTR priority distribution, resulting in more accurate and concise responses. Even with limited resources, using a small 
$k$ still achieves remarkable gains compared to the best baseline.
\begin{table}[h!tbp]
    \centering
    \small
    \setlength{\tabcolsep}{8pt}
    \renewcommand{\arraystretch}{0.78} 
        \begin{tabular}{lllll}
            \toprule
            \textbf{Metric} 
            & setting 
            & I.D. 
            & LP 
            & NC 
 \\
            \midrule

            & $k=1$ & 62.4  & 70.6 & 54.2\\
            
        Acc$\uparrow$ & $k=5$ & \underline{66.2} & \underline{72.1} & \underline{57.8} \\
            
            & $k=10$ & \textbf{69.5} & \textbf{72.4} & \textbf{58.8}\\
            & Best Baseline & 61.8 & 69.7 & 52.0 \\
            \midrule

            &  $k=1$ &  286.2 & 198.2 & 196.6 \\
        Tok$\downarrow$ & $k=5$ & \underline{232.4} & \underline{181.6} & \underline{158.9}\\
            &  $k=10$ & \textbf{212.0} & \textbf{177.6} & \textbf{136.1} \\
            & Best Baseline & 319.8 & 206.8 & 214.8\\


            \bottomrule
        \end{tabular}
        \vspace{-5pt}
        \caption{ The accuracy and token consumption with setting different $k$ values. The base VLM is GPT-4o.}
        \label{tab: k}
        \vspace{-5pt}
\end{table}

\subsection{Router Transferability for VLMs}
\begin{table}[htbp]
    \centering
    \small
    \setlength{\tabcolsep}{1.8pt}
    \renewcommand{\arraystretch}{0.8} 
        \begin{tabular}{lcccc}
            \toprule
            \textbf{Metric} 
            & I.D. 
            & LP 
            & NC 
            & \textbf{Avg.} \\
            \midrule
            \multicolumn{5}{c}{\textit{Transfer Gemini-2.5 Pro's Router to GPT-4o}} \\
            \midrule
            
            $\Delta \text{Acc} \uparrow$
             & -0.3(+6.1) & -0.1(+3.2) & -1.4(+6.6) & -0.6(+5.3)  \\

            \midrule
            
            $\Delta \text{Tok} \downarrow$
             & -35.5(-123.4) & -9.2(-56.4) & -40.1(-119.5) & -28.3(-99.8)   \\
            \midrule
            \multicolumn{5}{c}{\textit{Transfer GPT-4o's Router to Gemini-2.5 Pro}} \\
            \midrule
            
            $\Delta \text{Acc} \uparrow$
            & +0.1(+1.9) & +0.5(+4.0) & -0.1(+9.6) & +0.2(+5.2)  \\
            \midrule
            
            $\Delta \text{Tok} \downarrow$     & +14.4(-216.2) & +26.9(-47.5) & +3.2(-115.6) & +14.8(-126.4) \\
            \bottomrule
        \end{tabular}
\vspace{-5pt}
\caption{Performance differences when a GTR router trained for one VLM is applied to another VLM (i.e., transfer).}
\label{tab: transfer}
\vspace{-15pt}
\end{table}

To evaluate the cross-model generalization of the proposed method, we conducted transfer experiments where the router trained for one VLM (either GPT-4o or Gemini-2.5 Pro) is directly used by the other without retraining. This setup tests whether GTR preferences learned from one model's behavior could be effectively leveraged by another. In Table \ref{tab: transfer}, the values outside parentheses are changes relative to using the router aligned with the target VLM (i.e., naive router). Values in parentheses are changes relative to the best non-DynamicGTR baseline. As indicated, although transferred routers can slightly underperform the native router in some cases (reflecting model-specific biases), they consistently deliver benefits over standard baselines in both accuracy and efficiency, demonstrating that learned GTR preferences transfer across VLMs.

\section{Conclusion}
We propose DynamicGTR, a novel framework that leverages the preferences of queries on diverse graph topology representations. With extensive experiments, DynamicGTR boosts VLM-based graph capabilities in both in-domain tasks and out-of-domain applications, showing strong generalizability across tasks, models, and domains.


{
    \small
    \bibliographystyle{ieeenat_fullname}
    \bibliography{main}
}
\clearpage
\setcounter{page}{1}
\maketitlesupplementary
\renewcommand{\thetable}{\Alph{table}}

\renewcommand{\thefigure}{\Alph{figure}}

\section*{A. GTR Generation}
In this section, we provide more detailed introductions for the GTRs in $\mathcal{R}_{ZS}$ and their generation approaches. As introduced, $\mathcal{R}_{ZS}$ contains 8 types of dedicated GTRs, where $V_{dot}, V_{neato}, V_{circo}, V_{fdp}$ and $V_{sfdp}$ are visual GTRs, and $ T_{set}, T_{list}$, and $T_{mat}$ are textual GTRs.  

\paragraph{More Details for Visual GTRs Generation}
These visual GTRs are generated by leveraging diverse layout algorithms to compute graph layouts displayed on the canvas, specifically $V_{dot}$, $V_{neato}$, $V_{circo}$, $V_{fdp}$, and $V_{sfdp}$. Their generation follows the methodologies outlined in \citep{wei2024gita} and \citep{wei2024rendering}, using the graph visualization tool Graphviz \citep{graphviz}. To ensure style consistency without unnecessary disturbances, we only vary the layout algorithms while fixing other configurations (white background, circular node shape (except for barpartite graph matching, we use rectangular/circular shape to distinguish the hosts/tasks), default edge thickness) in Graphviz during generation. Specifically, the five visual GTRs, each utilizing a distinct layout algorithm, are introduced as follows:
\begin{figure}[htbp]
\centering
\includegraphics[width=0.45\textwidth]{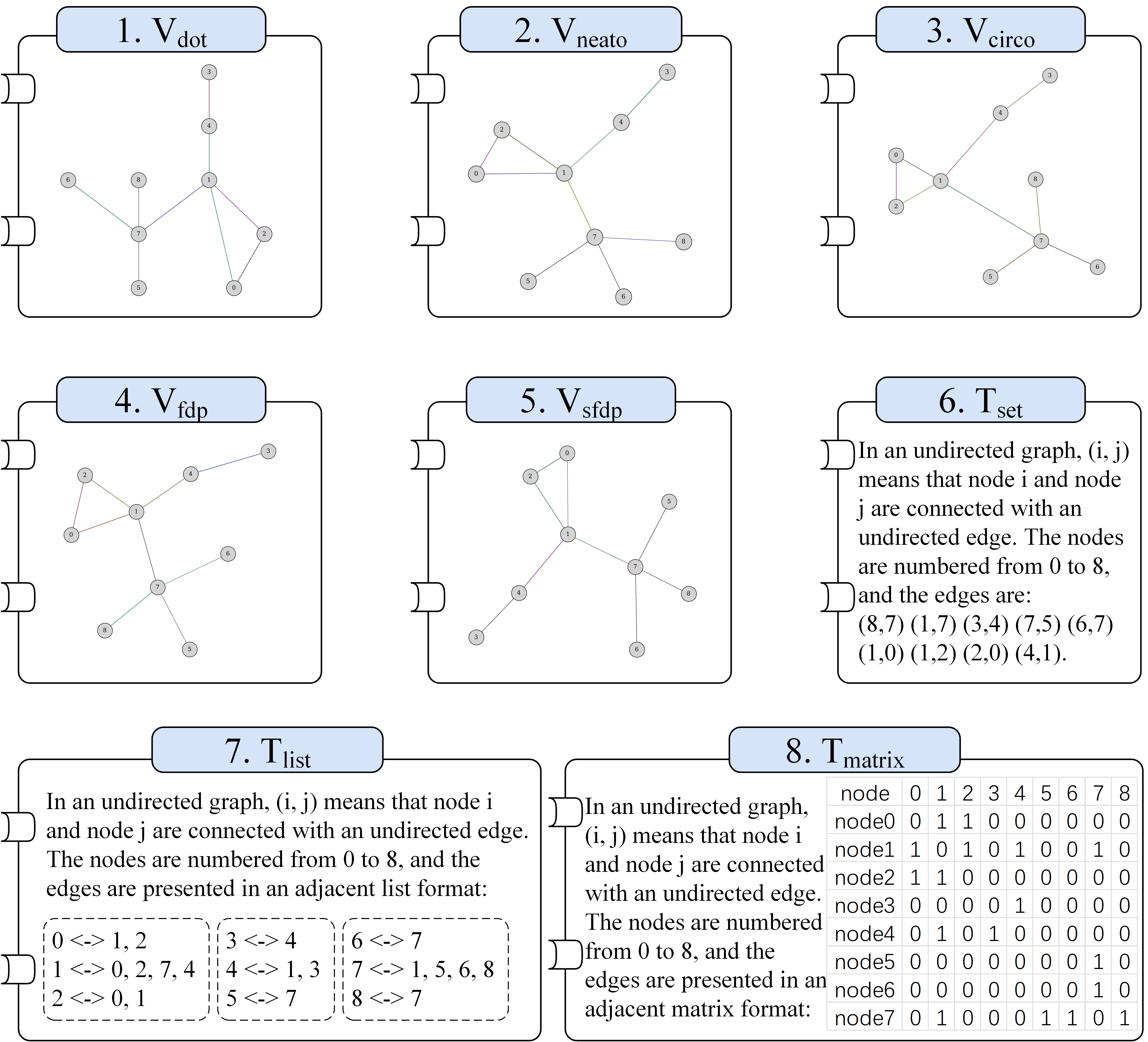}
\caption{An Illustration of eight candidate GTRs in $\mathcal{R}_{ZS}$.}
    \label{fig: examples2}
\end{figure}
\begin{itemize}
    \item \textbf{$V_{dot}$}: This visual GTR adopts a hierarchical layout algorithm, which minimizes edge crossings and maintains edge directions (either top-to-bottom or left-to-right). It is particularly effective for representing organizational structures or flowcharts as a visual topology.
    
    \item \textbf{$V_{neato}$}: As a visual GTR, it employs a spring model \citep{fruchterman1991graph} layout algorithm. By simulating edges as springs and nodes as mutually repelling entities, it generates an optimal arrangement through force simulation, making it well-suited for visualizing undirected graphs and network relationships.
    
    \item \textbf{$V_{circo}$}: This visual GTR uses a circular layout algorithm that places nodes in a ring structure to reduce edge crossings. It excels at highlighting cyclic topologies, such as those in electronic circuits or advanced abstract syntax trees.
    
    \item \textbf{$V_{fdp}$}: The visual GTR also relies on the spring model but with an optimized algorithm for large graphs. While prioritizing speed, it may sacrifice some node distribution uniformity compared to neato, balancing efficiency and visual clarity for larger topologies.
    
    \item \textbf{$V_{sfdp}$}: This visual GTR incorporates a multi-level force-directed algorithm to extend the fdp layout algorithm to even larger-scale graphs. It maintains force-directed layout characteristics while handling thousands of nodes efficiently, making it ideal for visualizing large topological structures.
\end{itemize}
To align with the scope of our paper, we do not delve further into additional algorithmic details of these layout methods. For comprehensive implementation specifics of each layout algorithm, readers are referred to the Graphviz technical paper \citep{graphviz} and its corresponding Python package documentation. Notably, we have integrated a dedicated interface within our codebase to facilitate the generation of these visual GTRs, and we recommend that users utilize this interface for direct invocation.

\paragraph{More Details for Textual GTRs Generation}
The textual $T_{set}$, $T_{list}$, and $T_{mat}$ present the graph topology structure via Edge set, Adjacent List, and Adjacent Matrix, respectively. Their prompt templates are detailed in Table \ref{tab: prompt1} to \ref{tab: prompt3}. For $T_{set}$, the prompts are derived from \citep{NLGraph}, while the prompts for $T_{list}$ and $T_{mat}$ are designed and polished by ourselves. To be specific, we introduce these textual GTRs as follows:

\begin{itemize}
    \item \textbf{$T_{set}$}: This textual GTR represents graph topology through an unordered collection of edge tuples, where each tuple explicitly denotes a connection between two nodes (e.g., $(u, v)$ for an edge from node $u$ to node $v$). By omitting redundant structural framing, it achieves high information density, storing all edge relationships in a compact format. This characteristic makes it particularly efficient for tasks requiring quick enumeration of all connections, such as edge existence checks or basic subgraph extraction. However, its unordered nature may complicate reasoning about node neighborhoods or hierarchical relationships, as no implicit grouping of edges by source node is provided.
    
    \item \textbf{$T_{list}$}: Structured as a node-centric inventory, this textual GTR organizes edges by their source nodes, listing all adjacent target nodes for each node in a sequentially ordered format (sorted by node labels). This design strengthens the visibility of neighbor relationships, as all connections originating from a specific node are grouped together, facilitating tasks like neighbor counting, path traversal, or community detection. The sorted arrangement of nodes and edges (by label) further enhances readability, enabling systematic scanning of topological patterns. Compared to $T_{set}$, it introduces moderate structural overhead but significantly improves accessibility to node-specific relationship information.
    
    \item \textbf{$T_{mat}$}: This textual GTR encodes graph topology as a tabular matrix where rows and columns correspond to nodes, and cell values indicate edge presence (e.g., 1 for an existing edge, 0 for absence). While the inclusion of 0 values for non-edges introduces redundancy, especially in sparse graphs, where most cells are 0, it offers exceptional intuitiveness for visualizing global connectivity patterns. The matrix structure allows for immediate identification of symmetric relationships (via diagonal symmetry) and dense subgraphs (via contiguous blocks of 1s), making it well-suited for tasks like bipartite graph analysis or adjacency pattern recognition. Despite its lower information density, the grid-based format aligns with human cognitive patterns for spatial relationship processing, reducing the cognitive load for certain types of topological reasoning.
\end{itemize}

\section*{B. GTRP Dataset Construction}
The GTRP dataset is designed to investigate query-specific preferences among GTRs within $\mathcal{R}_{ZS}$. To begin with, we generate 1K samples of QA for each task in \{Conn, Cyc, TP, SP, MF, BGM, HP\}, where the graph topologies are randomly generated by the Erdős–Rényi \citep{er_model} model ($\text{node count }N \in [3,30], \text{edge probability} \in [0.1,0.7]$). For SP and MF, the edge weights are randomly assigned $\in \{1,2,...,10\}$. Graphs that fail to meet task-specific validity criteria are regenerated (We make sure there exists at least a valid answer), ensuring we obtain 7K valid samples (1K per task).

Then, for each question, we evaluate every GTR in $\mathcal{R}_{ZS}$ by running $k$ independent trials to compute the corresponding GRE metric. GTRs are ranked for each question based on their GRE scores in descending order; in cases of tied GRE scores, the tied GTRs are treated as multi-label preferences. By pairing each question with its most preferred GTRs (those with the highest GRE scores), we construct the GTR Preference dataset, denoted as $\mathcal{D}_{\text{GTRP}} = \{(q_i, \mathcal{R}_{q_i}^*)\}$.

To validate response correctness, we use dedicated algorithms for each task:
\begin{itemize}
    \item Connectivity (Conn) identification answers are judged using the Union-Find algorithm \citep{tarjan1975efficiency}.
    \item Cycle detection (Cyc) results are verified via Depth-First Search (DFS) \citep{hopcroft1973algorithm}.
    \item Topological sorting (TS) answers are checked by verifying edge direction constraints in the sequence \citep{kahn1962topological}.
    \item Shortest path (SP) results are validated using Dijkstra's algorithm \citep{dijkstra2022note}.
    \item Maximum flow (MF) answers are judged via the Edmonds-Karp algorithm \citep{edmonds1972theoretical}.
    \item Bipartite graph matching (BGM) results are verified using the Hopcroft-Karp algorithm \citep{hopcroft1973n}.
    \item Hamiltonian path (HP) answers are checked using a backtracking-based approach \citep{bellman1962dynamic}.
\end{itemize}

The construction of $\mathcal{D}_{\text{GTRP}}$ is contextualized by both the target LMM and the specific definition of the GRE metric. Consequently, distinct LMMs require unique instantiations of the GTRP dataset, while adjustments to the $\alpha$ parameter, which modulates the trade-off between accuracy and brevity in necessitate corresponding updates to $\mathcal{D}_{\text{GTRP}}$ to reflect the shifted weightings in the metric.

\begin{table*}[h!tbp]
    \centering
    \setlength{\tabcolsep}{3pt}
    \renewcommand{\arraystretch}{1.4} 
    \small 
    \begin{tabular}{llccccccc}
        \hline
        Model & GTRs & Conn & Cyc & TS & SP & MF & BGM & HP \\
        \hline
        \multirow{8}{*}{\makecell[c]{GPT-4o}} & 1st 
        &\cellcolor{gray!20}$V_{fdp}$ (92.3\%) 
        &\cellcolor{gray!20} $V_{sfdp}$ (85.1\%)
        & $T_{set}$ (58.5\%)
        & $T_{set}$ (19.5\%)
        & $T_{list}$ (21.7\%)
        &\cellcolor{gray!20} $V_{dot}$ (34.2\%)
        &$T_{list}$ (30.4\%)\\
        \cline{2-9}
        & 2nd & 
        \cellcolor{gray!20}$V_{neato}$ (92.1\%) 
        & \cellcolor{gray!20}$V_{fdp}$ (12.9\%)
        & $T_{list}$ (36.2\%)
        &\cellcolor{gray!20} $V_{neato}$ (18.7\%)
        & $T_{set}$ (20.8\%)
        &\cellcolor{gray!20} $V_{circo}$ (20.3\%)
        & $T_{set}$ (20.3\%)\\ 
        \cline{2-9}
        & 3rd 
        &\cellcolor{gray!20} $V_{sfdp}$ (91.7\%)
        &\cellcolor{gray!20} $V_{dot}$ (12.2\%)
        &\cellcolor{gray!20} $V_{dot}$ (23.1\%)& $T_{list}$ (17.1\%)& $T_{mat}$ (16.7\%)
        &\cellcolor{gray!20} $V_{neato}$ (19.8\%)
        &\cellcolor{gray!20} $V_{circo}$ (18.8\%)\\

        \cline{2-9} & 4th 
        &\cellcolor{gray!20}$V_{circo}$ (84.4\%) 
        &\cellcolor{gray!20} $V_{circo}$ (10.5\%)
        & $T_{mat}$ (0.8\%)
        & \cellcolor{gray!20}$V_{fdp}$ (16.3\%)
        & \cellcolor{gray!20}$V_{neato}$ (10.8\%)
        &\cellcolor{gray!20} $V_{sfdp}$ (13.4\%)
        &\cellcolor{gray!20}$V_{dot}$ (17.4\%)\\
        \cline{2-9}
        & 5th & 
        \cellcolor{gray!20}$V_{dot}$ (73.0\%) 
        & \cellcolor{gray!20}$V_{fdp}$ (61.1\%)
        & \cellcolor{gray!20}$V_{sfdp}$ (0.8\%)
        &\cellcolor{gray!20} $V_{circo}$ (13.8\%)
        & \cellcolor{gray!20}$V_{fdp}$ (8.3\%)
        &\cellcolor{gray!20} $V_{fdp}$ (9.1\%)
        & \cellcolor{gray!20}$V_{sfdp}$ (15.9\%)\\ 
        \cline{2-9}
        & 6th 
        & $T_{list}$ (27.0\%)
        & $T_{list}$ (10.8\%)
        &\cellcolor{gray!20} $V_{neato}$ (0.8\%)& \cellcolor{gray!20}$V_{sfdp}$ (12.2\%)& \cellcolor{gray!20}$V_{dot}$ (7.5\%)
        &$T_{list}$ (7.0\%)
        &\cellcolor{gray!20} $V_{neato}$ (14.5\%)\\
        \cline{2-9} & 7th 
        &$T_{mat}$ (58.0\%) 
        &$T_{mat}$ (0.2\%)
        & \cellcolor{gray!20} $V_{circo}$ (0.8\%) 
        & \cellcolor{gray!20}$V_{dot}$ (10.6\%)
        & \cellcolor{gray!20}$V_{circo}$ (7.5\%)
        &  $T_{set}$ (1.6\%)
        & $T_{mat}$ (7.2\%)
        \\
        \cline{2-9}
        & 8th & $T_{set}$ (47.7\%) 
        & $T_{set}$ (0.0\%) 
        & \cellcolor{gray!20} $V_{fdp}$ (0.8\%)
        & $T_{mat}$ (6.5\%)
        & \cellcolor{gray!20}$V_{sfdp}$ (6.7\%)

        & $T_{mat}$ (0.5\%)
        & \cellcolor{gray!20}$V_{fdp}$ (0.0\%)\\ 
        \cline{2-9}
        \hline
        \multirow{8}{*}{\makecell[c]{Gemini \\ 2.5 Pro}} & 1st 
        &\cellcolor{gray!20} $V_{neato}$ (88.8\%)
        &\cellcolor{gray!20}
        $V_{neato}$ (97.2\%)
        &
        $T_{set}$ (41.0\%)
        &
        $T_{list}$ (48.6\%)
        &
        $T_{mat}$ (40.0\%)
        &\cellcolor{gray!20}
        $V_{fdp}$ (24.4\%)
        &
        $T_{list}$ (42.9\%)\\
        \cline{2-9}
        & 2nd &\cellcolor{gray!20}
        $V_{fdp}$ (88.2\%)
        & \cellcolor{gray!20} $V_{sfdp}$ (93.0\%)
        & $T_{list}$ (30.3\%)
        & $T_{mat}$ (37.6\%)
        & $T_{list}$ (36.0\%)
        & \cellcolor{gray!20} $V_{sfdp}$ (14.5\%)
        & $T_{set}$ (31.7\%)
        \\
        \cline{2-9}
        & 3rd 
        &\cellcolor{gray!20} $V_{sfdp}$ (88.2\%)
        & \cellcolor{gray!20} $V_{dot}$ (71.8\%)
        & $T_{mat}$ (15.4\%)
        & $T_{set}$ (26.6\%)
        & $T_{set}$ (14.0\%)
        & \cellcolor{gray!20} $V_{neato}$ (14.5\%)
        & $T_{mat}$ (30.2\%)\\
        \cline{2-9} 
        \cline{2-9} & 4th 
        &\cellcolor{gray!20}$V_{circo}$ (87.0\%) 
        &\cellcolor{gray!20} $V_{circo}$ (71.1\%)
        & \cellcolor{gray!20}$V_{dot}$ (6.9\%)
        & \cellcolor{gray!20}$V_{sfdp}$ (13.8\%)
        & \cellcolor{gray!20}$V_{sfdp}$ (6.0\%)
        &\cellcolor{gray!20} $V_{dot}$ (13.7\%)
        &\cellcolor{gray!20}$V_{fdp}$ (9.5\%)\\
        \cline{2-9}
        & 5th & 
        \cellcolor{gray!20}$V_{dot}$ (83.4\%) 
        & \cellcolor{gray!20}$V_{neato}$ (8.3\%)
        & \cellcolor{gray!20}$V_{fdp}$ (4.3\%)
        &\cellcolor{gray!20} $V_{dot}$ (11.0\%)
        & \cellcolor{gray!20}$V_{neato}$ (2.0\%)
        &$T_{mat}$ (12.2\%)
        & \cellcolor{gray!20}$V_{circo}$ (9.5\%)\\ 
        \cline{2-9}
        & 6th 
        & $T_{list}$ (61.3\%)
        & $T_{list}$ (0.5\%)
        &\cellcolor{gray!20} $V_{neato}$ (2.1\%)& \cellcolor{gray!20}$V_{fdp}$ (8.3\%)& \cellcolor{gray!20}$V_{circo}$ (2.0\%)
        & \cellcolor{gray!20}$V_{circo}$ (9.2\%)
        &\cellcolor{gray!20} $V_{sfdp}$ (9.6\%)\\
        \cline{2-9} & 7th 
        &$T_{mat}$ (8.5\%) 
        &$T_{mat}$ (5.2\%)
        & \cellcolor{gray!20} $V_{sfdp}$ (0.0\%) 
        & \cellcolor{gray!20}$V_{circo}$ (4.6\%)
        & \cellcolor{gray!20}$V_{dot}$ (0.0\%)
        &  $T_{set}$ (6.9\%)
        & $V_{dot}$ (0.0\%)
        \\
        \cline{2-9}
        & 8th & $T_{set}$ (5.1\%) 
        & $T_{set}$ (2.1\%) 
        & \cellcolor{gray!20} $V_{fdp}$ (0.8\%)
        & \cellcolor{gray!20}$V_{neato}$ (2.8\%)
        & \cellcolor{gray!20}$V_{fdp}$ (0.0\%)

        & $T_{list}$ (5.3\%)
        & \cellcolor{gray!20}$V_{fneato}$ (0.0\%)\\ 
        \cline{2-9}
        \hline
    \end{tabular}
    \caption{Complete ranking of GTRs with respect to their label frequency in the GTRP dataset of GPT-4o and Gemini-2.5 Pro. By differing visual or textual GTRs with colors, \textbf{the special preference patterns of tasks are explicitly exposed}.}
    \label{tab: preference_4o}
\end{table*}
\section*{C. Label Distributions in GTRP datasets}
In this section, we extend Table 2 in the main manuscript text to present the complete label distributions of GTRP datasets (Table \ref{tab: preference_4o}). According to this table, we have additional findings beyond those mentioned in the main body of the paper:
\begin{itemize}
    \item 1) When comparing GPT-4o and Gemini-2.5 Pro, they exhibit highly consistent task-specific GTR rankings, indicating the task-specific GTR preferences can go beyond model biases.
    \item 2) Specifically, for Conn and Cyc, the suboptimal textual GTRs $(T_{list}, T_{mat}, T_{set})$ also have a consistent inner priority across both models, with $T_{list} > T_{mat} > T_{set}$, highlighting the relative priority of textual GTRs in expressing topological connectivity effectively.
    \item 3) In the case of GPT-4o, even in tasks where other textual GTRs like $T_{list}$ and $T_{set}$ are preferred, $T_{mat}$ may be suboptimal, indicating a model-specific misalignment with this GTR form. However, this discrepancy is not observed in Gemini-2.5 Pro, suggesting that when someone intends to use $T_{mat}$ with its more intuitive graph adjacency representation, it is preferable to use Gemini-2.5 Pro over GPT-4o.
    \item 4) Analysis of the frequency gaps among the GTRs reveals that the preferences in Conn and Cyc are more pronounced, with larger variations in frequency. This is followed by BGM, and then TS, SP, MF, and HP, where the frequency gaps are more moderate.   
\end{itemize}

\section*{D. Data and Task Details}
\subsection*{1. Task Introduction}
We introduce Tasks in this section. The 7 in-domain graph algorithmic QA tasks include:
\begin{itemize}
\item \textbf{Connectivity}~\citep{sedgewick2001algorithms} (abbreviated as Conn): Assess whether two randomly chosen nodes $u$ and $v$ in an undirected graph are linked.

\item \textbf{Cycle}~\citep{sedgewick2001algorithms} (abbreviated as Cyc): Determine if there is a cycle present within an undirected graph.

\item \textbf{Topological Sort}~\citep{kahn1962topological} (abbreviated as TS): Identify a valid topological ordering for a directed acyclic graph. This sort provides a sequence of nodes such that for every directed edge $u \leftarrow v$, node $u$ precedes node $v$ in the sequence.

\item \textbf{Shortest Path}~\citep{dijkstra2022note} (abbreviated as SP): Locate the shortest route between two nodes in a weighted undirected graph. The shortest path is defined as the route with the smallest total edge weight connecting the two nodes.

\item \textbf{Maximum Flow}~\citep{ford1956maximal} (abbreviated as MF): Compute the maximum flow from a source node to a destination node in a network graph.

\item \textbf{Bipartite Graph Matching}~\citep{karp1990optimal} (abbreviated as BGM): Identify the largest matching set in a bipartite graph. A matching set consists of edges where no two edges share a common node.

\item \textbf{Hamilton Path}~\citep{gould2003advances} (abbreviated as HP): Discover a Hamiltonian path in an undirected graph. This path visits each node exactly once.

\end{itemize}
The 2 out-of-domain downstream application tasks are:
\begin{itemize}
    \item \textbf{Link Prediction}: Predict whether a link or edge exists between two nodes in a network, based on the current structure and attributes of the graph. It is one of the cornerstone tasks in graph learning and has wide applications in social network analysis \citep{social1}, recommendation systems \citep{lightgcn}, and biological network studies \citep{drug1}.

    \item \textbf{Node Classification}: This task is concerned with predicting the categories of nodes within a graph.  Node classification is widely used in applications such as community detection \citep{newman2004finding}, fraud detection \citep{akoglu2015graph}, and identifying roles in social networks \citep{wasserman1994social}.
\end{itemize}
\subsection*{2. Dataset Statistics}
We adopt the GVLQA-BASE \citep{wei2024gita} benchmark for in-domain tasks.  For out-of-domain applications, we adopt the ca-GrQC, ca-HepTh \citep{cadataset}, and 15 K queries from the large-scale dataset ogbl-PPA~\cite{ogb} for link prediction (LP), and use the PolBlog \citep{adamic2005political}, Cora \citep{yang2016revisiting}, and 15K queries from the large-scale ogbn-product\cite{ogb} dataset for the node classification (NC) task. Data statistics are provided in Table \ref{tab: data1} and \ref{tab: data2}.

\begin{table*}[!htbp]
\centering

\begin{tabular}{lccccccc}
\toprule
 & Conn & Cyc & TS & SP & MF & BGM & HP \\ 
\midrule
\#sample & 16,410 &  4,100 &  2,910 &  1,560 &  1,500 & 1,860 & 900 \\
\#nodes & 25.01 & 23.42 & 21.86 & 13.65 & 13.90 & 21.13 & 13.24 \\ 
\#edges & 95.46 & 23.66 & 114.10 & 23.99 & 49.16 & 51.03 & 45.05 \\ \bottomrule
\end{tabular}
\caption{Data Statistics for in-domain graph algorithmic QA tasks.}
\label{tab: data1}
\end{table*}

\begin{table*}[!hbtp]
\centering
\resizebox{0.9\textwidth}{!}{%
    \begin{tabular}{ccccccc}
    \toprule
     & ca-GrQC & ca-HepTh & ogbl-ppa & PolBlogs & Cora & ogbn-product \\ 
     \midrule
    \# Nodes & 5,242 & 9,877 & 576,289 & 1,490 & 2,708 & 2,449,029 \\
    \# Edges & 14,496 & 25,998 & 30,326,273 & 19,025 & 5,278 & 61,859,140 \\
    domain  & collaboration  & collaboration & protein association & social & citation  & e-commerce \\
    average degree & 5.53 & 5.26 & 105.25 & 25.54 & 3.9 & 50.52 \\
    \bottomrule
    \end{tabular}%
}
\caption{Dataset Statistics for out-of-domain real-world applications}
\label{tab: data2}
\end{table*}

\section*{E. Results for LLaVA-OneVision-7B and Qwen3-VL-8B}
An advantage of the DynamicGTR framework is its generally usability on closed-source VLMs, because it does not require any access to model parameter or architecture details, but that not means DynamicGTR cannot run on open-source VLMs. We have shown its significant effectiveness on leading closed-source VLMs including GPT-4o and Gemini-2.5-Pro in the main body of the manuscript. Here in this supplementary material section, we supplement results on the leading open-source VLMs, including LLaVA-OneVision-7B \cite{li2024llavaonevision} and Qwen3-VL-8B \cite{qwen3} in Table \ref{tab:more_main1} and \ref{tab:more_main2}.

\begin{table*}[h!tbp]
    \centering
    \setlength{\tabcolsep}{1.5pt}
    \renewcommand{\arraystretch}{0.9} 
\small 
\begin{tabular}{cccccccccccccccc}
    \hline
    & \multicolumn{2}{c}{\raisebox{-0.9ex}[0pt]{Conn}} 
    & \multicolumn{2}{c}{\raisebox{-0.9ex}[0pt]{Cyc}} 
    & \multicolumn{2}{c}{\raisebox{-0.9ex}[0pt]{TS}} 
    & \multicolumn{2}{c}{\raisebox{-0.9ex}[0pt]{SP}} 
    & \multicolumn{2}{c}{\raisebox{-0.9ex}[0pt]{MF}} 
    & \multicolumn{2}{c}{\raisebox{-0.9ex}[0pt]{BGM}} 
    & \multicolumn{2}{c}{\raisebox{-0.9ex}[0pt]{HP}} \\
    Method
    &\multicolumn{2}{c}{\raisebox{0.5ex}{\rule{2.05cm}{0.4pt}}}
    &\multicolumn{2}{c}{\raisebox{0.5ex}{\rule{2.05cm}{0.4pt}}}
    &\multicolumn{2}{c}{\raisebox{0.5ex}{\rule{2.05cm}{0.4pt}}}
    &\multicolumn{2}{c}{\raisebox{0.5ex}{\rule{2.05cm}{0.4pt}}}
    &\multicolumn{2}{c}{\raisebox{0.5ex}{\rule{2.05cm}{0.4pt}}}
    &\multicolumn{2}{c}{\raisebox{0.5ex}{\rule{2.05cm}{0.4pt}}}
    &\multicolumn{2}{c}{\raisebox{0.5ex}{\rule{2.05cm}{0.4pt}}}
    \\
     & Acc & Tok & Acc & Tok & Acc & Tok & Acc & Tok & Acc & Tok & Acc & Tok & Acc & Tok \\
    \midrule
    \multicolumn{15}{c}{\textit{LLaVA-OneVision-7B}} \\
    \midrule
    CoT
    & 85.2 & \underline{312.5}
    & 45.3 & 528.8
    & 30.1 & 268.4
    & 48.2 & 612.7
    & 18.7 & 405.3
    & 60.3 & 412.8
    & 42.5 & \underline{156.3}
    \\
    NLGraph
    & 85.7 & 338.9
    & 52.1 & 382.4
    & 29.8 & \underline{241.7}
    & 52.6 & 586.2
    & 19.2 & \textbf{378.5}
    & 54.8 & \underline{398.6}
    & 48.9 & 212.7
    \\
    GraphDPR
    & \underline{88.6} & 456.3
    & \underline{60.5} & 689.2
    & 34.7 & 458.9
    & \underline{53.9} & \underline{547.8}
    & \underline{25.3} & 556.2
    & 68.9 & 635.7
    & \underline{55.2} & 521.4
    \\
    GITA
    & 86.9 & 329.7
    & 56.8 & \underline{372.1}
    & \textbf{35.2} & 301.4
    & 50.3 & \textbf{536.9}
    & 20.9 & \underline{408.2}
    & \underline{74.2} & 438.9
    & 47.6 & 235.1
    \\
    \rowcolor{gray!20} 
    DynamicGTR
    & \textbf{92.3} & \textbf{52.4}
    & \textbf{81.6} & \textbf{92.7}
    & \underline{34.9} & \textbf{207.5}
    & \textbf{62.8} & 551.3
    & \textbf{30.8} & 421.7
    & \textbf{85.7} & \textbf{276.3}
    & \textbf{58.6} & \textbf{98.2}
    \\
    \midrule
    \multicolumn{15}{c}{\textit{Qwen3-VL-8B}} \\
    \midrule
    CoT
    & 87.4 & \underline{295.8}
    & 48.6 & 502.3
    & 32.5 & 251.8
    & 50.7 & 589.4
    & 21.5 & 389.6
    & 63.7 & 395.2
    & 45.8 & \underline{142.7}
    \\
    NLGraph
    & 87.9 & 319.2
    & 55.3 & 368.7
    & 32.1 & \underline{229.4}
    & 55.2 & 561.8
    & 22.1 & \textbf{362.8}
    & 58.3 & \underline{381.4}
    & 51.6 & 198.3
    \\
    GraphDPR
    & \underline{90.2} & 438.7
    & \underline{63.2} & 657.9
    & 36.8 & 432.5
    & \underline{56.4} & \underline{523.6}
    & \underline{27.5} & 531.9
    & 71.4 & 602.8
    & \underline{57.9} & 498.6
    \\
    GITA
    & 89.1 & 312.4
    & 60.1 & \underline{356.8}
    & \textbf{37.5} & 284.6
    & 53.1 & \textbf{512.4}
    & 23.8 & \underline{392.5}
    & \underline{77.6} & 419.3
    & 50.3 & 218.5
    \\
    \rowcolor{gray!20} 
    DynamicGTR
    & \textbf{94.5} & \textbf{45.6}
    & \textbf{84.1} & \textbf{85.3}
    & \underline{37.2} & \textbf{192.8}
    & \textbf{65.2} & 528.7
    & \textbf{33.4} & 405.9
    & \textbf{88.3} & \textbf{259.7}
    & \textbf{61.4} & \textbf{89.5}
    \\
    \hline
\end{tabular}

    \caption{Zero-shot capabilities on in-domain graph algorithmic tasks for LLaVA-OneVision-7B and Qwen3-VL-8B. Acc and Tok refer to task-average accuracy (\%) and token consumption, respectively.}
    \label{tab:more_main1}
\end{table*}

\begin{table*}[tbp]
    \centering
    \setlength{\tabcolsep}{5.5pt}
    \renewcommand{\arraystretch}{1.0} 
\small 
\begin{tabular}{lcccccccccccc}
    \hline
    & \multicolumn{2}{c}{\raisebox{-0.9ex}[0pt]{ca-GrQC}} 
    & \multicolumn{2}{c}{\raisebox{-0.9ex}[0pt]{ca-HepTh}} 
    & \multicolumn{2}{c}{\raisebox{-0.9ex}[0pt]{ogbl-ppa}} 
    & \multicolumn{2}{c}{\raisebox{-0.9ex}[0pt]{PolBlog}} 
    & \multicolumn{2}{c}{\raisebox{-0.9ex}[0pt]{Cora}} 
    & \multicolumn{2}{c}{\raisebox{-0.9ex}[0pt]{ogbn-product}}  \\
    Method
    &\multicolumn{2}{c}{\raisebox{0.5ex}{\rule{2.05cm}{0.4pt}}}
    &\multicolumn{2}{c}{\raisebox{0.5ex}{\rule{2.05cm}{0.4pt}}}
    &\multicolumn{2}{c}{\raisebox{0.5ex}{\rule{2.05cm}{0.4pt}}}
    &\multicolumn{2}{c}{\raisebox{0.5ex}{\rule{2.05cm}{0.4pt}}}
    &\multicolumn{2}{c}{\raisebox{0.5ex}{\rule{2.05cm}{0.4pt}}}
    & \multicolumn{2}{c}{\raisebox{0.5ex}{\rule{2.05cm}{0.4pt}}}
    \\
     & Acc & Tok & Acc & Tok & Acc & Tok & Acc & Tok & Acc & Tok & Acc & Tok \\
    \midrule
    LLaVA-OneVision-7B
    & 58.4 & \underline{229.7}
    & 69.2 & \underline{276.3}
    & 34.8 & \underline{241.8}
    & 51.5 & 298.4
    & 44.7 & 265.9
    & 26.8 & 235.2
    \\
     \ +CoT
     & 60.9 & 278.5
     & 69.7 & 312.8
     & 40.3 & 285.4
     & 52.8 & 356.7
     & 46.9 & 312.5
     & 26.7 & 243.8
    \\
     \ +GITA
     & \underline{65.7} & 245.3
     & \underline{75.8} & 298.6
     & \underline{45.2} & 312.9
     & \underline{56.2} & \underline{269.3}
     & \underline{49.2} & \underline{221.7}
     & \underline{29.3} & 278.6
    \\
     \rowcolor{gray!20}\ +DynamicGTR
     & \textbf{69.3} & \textbf{172.5}
     & \textbf{79.5} & \textbf{228.4}
     & \textbf{47.6} & \textbf{253.7}
     & \textbf{63.8} & \textbf{189.6}
     & \textbf{56.3} & \textbf{128.9}
     & \textbf{33.5} & \textbf{192.4}
    \\
    \midrule
    Qwen3-VL-8B
    & 60.7 & \underline{215.3}
    & 71.5 & \underline{261.8}
    & 36.5 & \underline{228.5}
    & 53.8 & 282.6
    & 46.9 & 251.3
    & 28.4 & 221.7
    \\
     \ +CoT
     & 63.2 & 262.8
     & 72.0 & 298.5
     & 42.1 & 271.9
     & 55.1 & 338.9
     & 49.2 & 297.8
     & 28.3 & 230.5
    \\
     \ +GITA
     & \underline{67.9} & 232.6
     & \underline{77.3} & 285.4
     & \underline{46.8} & 297.6
     & \underline{58.5} & \underline{254.8}
     & \underline{51.6} & \underline{205.2}
     & \underline{31.1} & 265.3
    \\
     \rowcolor{gray!20}\ +DynamicGTR
     & \textbf{71.8} & \textbf{159.8}
     & \textbf{81.6} & \textbf{215.7}
     & \textbf{49.3} & \textbf{236.2}
     & \textbf{66.4} & \textbf{175.2}
     & \textbf{58.8} & \textbf{112.6}
     & \textbf{35.7} & \textbf{178.9}
    \\
    \hline
\end{tabular}
    \caption{Zero-shot capacities on out-of-domain real-world applications for LLaVA-OneVision-7B and Qwen3-VL-8B. Acc and Tok refer to task-average accuracy (\%) and token consumption, respectively.}
    \label{tab:more_main2}
    \vspace{-15pt}
\end{table*}
\section*{F. Implementation Details}
In our experiment, we use DeBERTaV3-base as GTR Router, which is trained with learning rate $\in \{5e-5, 5e-6\}$, weight decay $\in \{1e-2, 1e-3\}$, epoch $\in \{6, 8, 10\}$, batch size $\in \{16, 32, 64\}$. All experiments are conducted on a single NVIDIA A100 GPU. For all methods in comparison, the task instructions are unified in Table \ref{tab: instruction}, which is followed by a control instruction (provided in \ref{tab: control}) to specify the place and format of the answer to be extracted. More implementation details for each method are as follows:
\begin{itemize}
    \item GPT-4o/Gemini-2.5 Pro: For the backbone LMMs, we examine their graph QA performances by invoking the official API Interface to directly answer the questions with instructions. Due to the prevalent nature of edge set in existing works, we adopt $T_{set}$ to represent the graph topology during implementation.
    \item Vanilla Chain-of-thought (CoT) \citep{wei2022chain}: CoT has demonstrated that the simple prompts that drive the LMMs in analytical, step-by-step thinking can boost the general range of capabilities of LMM reasoning. We adopt CoT in our baseline by appending a sentence of CoT prompt in the input tail: ``\texttt{Please think step by step and present the rationales in a well-structured manner, to make the answer more reliable and robust.}''. Due to the prevalent nature of edge set in existing works, we adopt $T_{set}$ to represent the graph topology during implementation.
    \item NLGraph \citep{NLGraph}/ GraphDPR \citep{li2024visiongraph}/ GITA \citep{wei2024gita}: NLGraph proposes to use BAG prompting to conceptualize a graph and algorithmic prompting to specify the algorithm. GraphDPR proposes to employ the external graph algorithmic toolkits to generate intermediate descriptions and code, which are utilized to enhance multi-step reasoning. GITA simultaneously adopts a visual GTR and (similar to $V_{dot}$) a textual GTR (Similar to $T_{set}$) to solve graph algorithmic problems by leveraging the complementary nature of the visual and textual mutual-enhancement across tasks. We evaluate these methods based on their own official implementations, but modifying the prompt templates to that we have specified in Table \ref{tab: prompt1}, \ref{tab: instruction}, and \ref{tab: control}.
\end{itemize}

\begin{table*}[htbp]
  \centering 
	\renewcommand{\arraystretch}{1.5}
\resizebox{\linewidth}{!}{
  \begin{tabular}
  {p{0.8in}|p{6.2in}} \toprule
 \textbf{Tasks}
		&  \textbf{Task Instruction}  \\
		\midrule
		Connectivity
		& \texttt{Is there a path between node [P] and node [P] in this undirected graph?} 
	\\ \cline{1-2}
    Cycle
		& \texttt{Is there a cycle in this undirected graph?} 
	\\ \cline{1-2}

Topological Sort & \texttt{This representation depicts a directed graph, in which each directed edge from node A to node B signifies that, according to the topological order, node A must precede node B. Q: The topological order of the directed graph is:} 
	\\ \cline{1-2}
Shortest Path & \texttt{This representation illustrates a directed graph, with each edge's capacity indicated by a numerical label in close proximity.Q: What is the maximum flow from node 4 to node 0:} 
 \\ \cline{1-2} 
Maximum Flow & \texttt{This representation illustrates a directed graph, with each edge's capacity indicated by a numerical label in close proximity. Q: What is the maximum flow from node [P] to node [P]:}
\\ \cline{1-2}
Bipartite Graph Matching & \texttt{There are [P] hosts numbered from [P] to [P], and [P] tasks numbered from [P] to [P]. Each host has a set of tasks that it is interested in, represented by arrows from a host to a task in the diagram. However, each host is capable of solving only one task, and similarly, each task can be resolved by just one host. Q:  What is the maximum number of hosts that can be assigned a task they are interested in?} 
 \\  \hline
Hamilton Path & \texttt{Q: Begin with node 0, what is the path in this graph that visits every node exactly once?} \\
\hline
 Link Predict
		& \texttt{The task is link prediction, aiming to predict the presence or absence of an unknown edge between Node [P] and Node [P] based on the known graph structure. Q: Does an unknown edge exist between Node [P] and Node [P]?}\\

\hline
 Node Classify
		& \texttt{The task is semi-supervised node classification, and needs to predict which class Node [P] belongs to, based on graph structure and known node classes. Q: Node [P] belongs to Class: Note that capacity is directional, allowing flow only in the edge direction; reverse edge direction should not be considered in the path. }\\
\hline
  \end{tabular}}
    \caption{Unified Task Instructions in experimental comparisons, where [P]s are placeholders that will be substituted in specific questions.
    }
    \label{tab: instruction}
  \end{table*}

\begin{table*}[htbp]
  \centering 
	\renewcommand{\arraystretch}{1.1}
\resizebox{\linewidth}{!}{
  \begin{tabular}
  {p{0.8in}|p{6.2in}} \toprule
 \textbf{Tasks}
		&  \textbf{Task Instruction}  \\
		\midrule
		Connectivity/\newline Cycle/\newline Link Predict
		& \texttt{Please put the answer between <answer> and </answer> tags. For example, <answer>Yes</answer> or <answer>No</answer>.} 
	\\ \cline{1-2}

Topological Sort/\newline Shortest Path/\newline Hamilton Path& \texttt{Please put the answer between <answer> and </answer> tags. For example, <answer>0->1->2->3->4</answer> or <answer>0->1->3->7->8->4->6->5->9->2</answer>. } 
	\\ \cline{1-2}
Maximum Flow & \texttt{Please put the answer between <answer> and </answer> tags. For example, <answer>3</answer> or <answer>8</answer>. } 
 \\  \hline
Node Classify & \texttt{Please put the answer between <answer> and </answer> tags. For example, <answer>Class 1</answer> or <answer>Class 3</answer>.} \\
\hline
  \end{tabular}}
    \caption{Control Task Instructions in experimental comparisons.
}
    \label{tab: control}
  \end{table*}

\section*{G. Ablation Study: More results on other VLMs}

To demonstrate the importance of the proposed GTR Routing, we present metrics comparing individual GTRs with the GTR Router within the DynamicGTR framework. 

While we have already presented such an analysis for GPT-4o in our manuscript in section 5.3 based on Table 5, it is important to note that Table 5 only contains results for GPT-4o. Here, we provide a complete comparison table including the results for more VLMs in Table \ref{tab: abl2}. The observations for other VLMs are similar to those we have reported in section 5.3 for GPT-4o, indicating that there is no single dominant GTR across all tasks. The top GTRs for each task align closely with the task preferences shown in the GTRP dataset (Table \ref{tab: preference_4o}).

With the GTR Router, DynamicGTR achieves the highest GRE scores across all tasks compared to individual GTRs, showcasing its superior performance in balancing accuracy and efficiency. 

\begin{table*}[hbpt]
    \centering
    \setlength{\tabcolsep}{1.5pt}
    \renewcommand{\arraystretch}{0.8} 
\small 
\begin{tabular}{lcccccccccccccccc}
    \hline
    &  & \multicolumn{2}{c}{\raisebox{-0.6ex}[0pt]{Conn}} 
    & \multicolumn{2}{c}{\raisebox{-0.6ex}[0pt]{Cyc}} 
    & \multicolumn{2}{c}{\raisebox{-0.6ex}[0pt]{TS}} 
    & \multicolumn{2}{c}{\raisebox{-0.6ex}[0pt]{SP}} 
    & \multicolumn{2}{c}{\raisebox{-0.6ex}[0pt]{MF}} 
    & \multicolumn{2}{c}{\raisebox{-0.6ex}[0pt]{BGM}} 
    & \multicolumn{2}{c}{\raisebox{-0.6ex}[0pt]{HP}} 
    & \\
    GTR & Avg.GRE
    &\multicolumn{2}{c}{\raisebox{0.5ex}{\rule{2.05cm}{0.4pt}}}
    &\multicolumn{2}{c}{\raisebox{0.5ex}{\rule{2.05cm}{0.4pt}}}
    &\multicolumn{2}{c}{\raisebox{0.5ex}{\rule{2.05cm}{0.4pt}}}
    &\multicolumn{2}{c}{\raisebox{0.5ex}{\rule{2.05cm}{0.4pt}}}
    &\multicolumn{2}{c}{\raisebox{0.5ex}{\rule{2.05cm}{0.4pt}}}
    &\multicolumn{2}{c}{\raisebox{0.5ex}{\rule{2.05cm}{0.4pt}}}
    &\multicolumn{2}{c}{\raisebox{0.5ex}{\rule{2.05cm}{0.4pt}}}
    \\
           &  & Acc & Tok & Acc & Tok & Acc & Tok & Acc & Tok & Acc & Tok & Acc & Tok & Acc & Tok \\
    \midrule
    \multicolumn{16}{c}{\textit{LLaVA-OneVision-7B}} \\
    \midrule
    $V_{dot}$ & 1.11
    &72.3 & 9.1
    &68.5 & \textbf{8.0}
    &10.1 & \textbf{35.2}
    &12.7 & \underline{168.5}
    &6.2 & 435.8
    &80.1 & \textbf{265.9}
    &11.5 & 45.2
    \\
    $V_{neato}$ & 1.09
    &88.6 & \textbf{7.9}
    &\underline{76.3} & \textbf{8.0}
    &2.8 & \textbf{35.2}
    &18.5 & \textbf{152.3}
    &9.5 & 408.7
    &\underline{83.4} & 281.7
    &9.8 & \textbf{42.7}
    \\
    $V_{circo}$ & 0.99
    &80.5 & 8.6
    &62.7 & \textbf{8.0}
    &2.5 & \textbf{35.2}
    &16.3 & 175.8
    &7.8 & 442.1
    &78.9 & \underline{278.4}
    &12.9 & \textbf{42.7}
    \\
    $V_{fdp}$ & 0.76
    &\underline{90.2} & \underline{8.4}
    &54.1 & \textbf{8.8}
    &2.5 & \textbf{35.2}
    &15.6 & 169.7
    &7.8 & 410.3
    &75.3 & 302.6
    &3.8 & 78.5
    \\
    $V_{sfdp}$ & 1.07
    &87.9 & \underline{8.4}
    &73.5 & \textbf{8.0}
    &6.1 & 142.8
    &22.8 & 186.4
    &7.8 & \underline{407.2}
    &77.6 & 310.2
    &10.6 & 45.2
    \\
    $T_{set}$ & 0.83
    &85.2 & 312.5
    &45.3 & 528.8
    &\underline{30.1} & 268.4
    &48.2 & 612.7
    &\underline{18.7} & \textbf{405.3}
    &60.3 & 412.8
    &\underline{42.5} & 156.3
    \\
    $T_{list}$ & 0.85
    &82.7 & 265.9
    &46.8 & 412.3
    &28.5 & 245.7
    &\underline{49.8} & 573.2
    &17.9 & 432.8
    &58.6 & 445.1
    &42.5 & 132.6
    \\
    $T_{mat}$ & 0.32
    &73.1 & 282.4
    &43.9 & 475.6
    &5.9 & 412.8
    &32.1 & 648.5
    &12.3 & 456.9
    &52.8 & 462.7
    &28.7 & 185.4
    \\
    \rowcolor{gray!20}  Router & \textbf{1.55}
    & \textbf{92.3} & 22.4
    & \textbf{81.6} & 92.7
    & \textbf{34.9} & 177.5
    & \textbf{62.8} & 551.3
    & \textbf{30.8} & 421.7
    & \textbf{85.7} & 282.3
    & \textbf{58.6} & 98.2
    \\
    \midrule
    \multicolumn{16}{c}{\textit{Qwen3-VL-8B}} \\
    \midrule
    $V_{dot}$ & 1.19
    &75.8 & 8.9
    &71.2 & \underline{10.4}
    &11.8 & 38.2
    &14.3 & 169.7
    &7.1 & 420.5
    &82.6 & \textbf{258.3}
    &12.8 & \textbf{41.3}
    \\
    $V_{neato}$ & 1.17
    &91.3 & \textbf{7.5}
    &78.6 & \underline{10.4}
    &3.2 & \textbf{33.8}
    &20.7 & \textbf{145.2}
    &10.9 & \underline{395.8}
    &\underline{86.1} & 272.5
    &10.5 & \textbf{41.3}
    \\
    $V_{circo}$ & 1.07
    &83.7 & 8.4
    &65.9 & \underline{10.4}
    &2.9 & \textbf{33.8}
    &17.8 & 168.4
    &8.9 & 428.7
    &80.5 & 269.1
    &13.6 & \textbf{41.3}
    \\
    $V_{fdp}$ & 0.85
    &\underline{92.5} & \underline{8.1}
    &57.3 & \underline{10.4}
    &2.9 & \textbf{33.8}
    &16.9 & \underline{162.3}
    &8.9 & 398.4
    &78.2 & 291.7
    &4.1 & 76.8
    \\
    $V_{sfdp}$ & 1.16
    &90.6 & \underline{8.1}
    &\underline{79.8} & \textbf{8.4}
    &6.7 & 138.5
    &24.2 & 179.6
    &8.9 & \textbf{392.6}
    &80.1 & 298.4
    &11.9 & 43.5
    \\
    $T_{set}$ & 0.92
    &87.4 & 295.8
    &48.6 & 502.3
    &\underline{32.5} & 251.8
    &50.7 & 589.4
    &\underline{21.5} & 398.6
    &63.7 & 395.2
    &\underline{45.8} & 142.7
    \\
    $T_{list}$ & 0.95
    &85.1 & 248.6
    &51.2 & 387.6
    &30.8 & 232.4
    &\underline{52.9} & 556.8
    &20.3 & 415.2
    &60.9 & 421.8
    &45.8 & 128.4
    \\
    $T_{mat}$ & 0.42
    &76.3 & 269.2
    &47.1 & 453.9
    &6.5 & 398.7
    &34.8 & 625.3
    &13.8 & 438.5
    &55.2 & 447.3
    &30.2 & 172.6
    \\
    \rowcolor{gray!20}  Router & \textbf{1.63}
    & \textbf{94.5} & \textbf{15.6}
    & \textbf{84.1} & 85.3
    & \textbf{37.2} & 192.8
    & \textbf{65.2} & 528.7
    & \textbf{33.4} & 405.9
    & \textbf{88.3} & \underline{266.7}
    & \textbf{61.4} & 89.5
    \\
    \midrule
    \multicolumn{16}{c}{\textit{GPT-4o}} \\
    \midrule
    $V_{dot}$ & 1.16
    &78.5 & 8.4
    &80.0 & \underline{8.0}
    &12.3 & 346.0
    &14.5&\underline{146.3}
    &7.5&410.2
    &\underline{91.2}&\underline{244.3}
    &13.3&\textbf{40.0}
    \\
    $V_{neato}$ & 1.20
    &95.1&\textbf{7.7}
    &\underline{87.1}&\underline{8.0}
    &3.0&\textbf{30.0}
    &20.0&\textbf{130.1}
    &10.8&387.4
    &87.2&280.0
    &11.1&\textbf{40.0}
    \\
    $V_{circo}$ & 1.13
    &89.7& 8.2
    &70.7&\underline{8.0}
    &3.0&\textbf{30.0}
    &18.2&153.5
    &8.1&421.3
    &88.2&271.6
    &14.4&\textbf{40.0}
    \\
    $V_{fdp}$ & 0.88
    &\underline{96.0}&\underline{8.1}
    &60.5&8.0
    &3.0&\textbf{30.0}
    &17.3&150.5
    &8.1&389.2
    &82.9&295.3
    &4.4&74.0
    \\
    $V_{sfdp}$ & 1.21
    &94.8&\underline{8.1}
    &85.1&\textbf{7.0}
    &7.0&120.0
    &25.5&168.7
    &8.1&386.1
    &84.0&302.9
    &12.2&40.0
    \\
    $T_{set}$ & 1.05
    & 92.5&273.3
    & 52.7&480.6
    & \underline{36.6}&224.2
    & 54.6&566.0
    & \underline{25.3}&\textbf{362.9}
    & 69.5&370.1
    & \underline{50.0}&124.9
    \\
    $T_{list}$ & 1.07
    & 89.0&218.2
    & 53.2&359.7
    & 34.2&206.3
    & \underline{55.5}&518.9
    & 24.2&400.7
    & 65.8&410.8
    & 50.0&107.0
    \\
    $T_{mat}$ & 0.56
    &79.9&233.7
    &52.7&417.1
    &6.8&378.1
    &34.5&591.8
    &17.2&402.9
    &60.4&407.3
    &31.1&160.9
    \\
    \rowcolor{gray!20}  Router & \textbf{1.64}
    & \textbf{96.1}&38.8 
    & \textbf{89.3}&75.9
    & \textbf{41.4}&176.1
    & \textbf{68.4}&499.1
    & \textbf{36.6}&\underline{385.2}
    & \textbf{92.0}&\textbf{233.6}
    & \textbf{61.1}&76.3
    \\
    \midrule
    \multicolumn{16}{c}{\textit{Gemini-2.5 Pro}} \\
    \midrule
    $V_{dot}$ &  0.98
    &94.1 & 8.4
    &72.5 & \textbf{8.0}
    &23.2 & 1157.0
    &46.8& 846.0
    &14.6& 1113.5
    &93.9&1023.7
    &30.6&1196.0
    \\
    $V_{neato}$ & 1.00
    & 99.6 & \textbf{7.7}
    &97.9&\textbf{8.0}
    &7.9&\underline{885.0}
    & 53.2&866.9
    &25.0&1173.3
    &96.5&1005.7
    &27.4&1056.0
    \\
    $V_{circo}$ & 0.85
    & 98.4&9.5 
    &71.1&\textbf{8.0}
    &3.0&1106.0
    &40.4&852.8
    &8.3&1056.4 
    &91.3&1137.7
    &226.0&\textbf{30.0}
    \\
    $V_{fdp}$ & 1.05
    &\underline{99.6}&9.7
    & 61.0 & \textbf{8.0}
    &6.1&\textbf{638.0}
    &44.0&856.8
    &18.8&1158.0
    &\underline{98.3}&\underline{948.2}
    &21.0&\textbf{30.0}\\
    
    $V_{sfdp}$ & 0.80
    &99.4&\underline{8.4}
    &93.7&\textbf{8.0}
    &5.5&1585.0
    &46.8&828.7
    &16.7&1178.7
    &95.7&1042.2
    &30.6&891.0
    \\
    $T_{set}$ & 1.23
    &97.2&218.7
    &\underline{98.6}&716.6
    &\underline{84.1}&1395.9
    &93.6&810.8
    &91.7&1154.9
    &97.1&1076.8
    &96.8&678.1
    \\
    $T_{list}$ & 1.27
    &97.4&155.3
    & 96.5&652.8
    & 76.8&1532.3
    & \textbf{96.3}&\textbf{601.0}
    & 92.2&\underline{1035.8}
    & 97.6&1129.9
    & 97.0&646.1
    \\
    $T_{mat}$ & 1.22
    &97.6&176.1
    &96.5&720.8
    &68.9&1554.7
    & 93.6&\underline{728.7}
    & \underline{95.8}&1042.6
    & 96.9&1033.8
    & \underline{98.4}&737.0\\
    \rowcolor{gray!20} Router & \textbf{1.85}
    &\textbf{100}&\textbf{12.9}
    &\textbf{99.3}&16.7
    &\textbf{87.8}&1191.2
    &\textbf{96.3}&798.6
    &\textbf{100}&\textbf{1004.8}
    &\textbf{100}&\textbf{776.0}
    &\textbf{100}&254.6
    \\
\hline
\end{tabular}
    \caption{Comparison of the DynamicGTR framework with GTR Router versus single GTRs on in-domain tasks across diverse leading VLMs.}
    \label{tab: abl2}
    \vspace{-5pt}
\end{table*}

\begin{table*}[htbp]
  \footnotesize 
  \centering 
	\renewcommand{\arraystretch}{1.1}
\resizebox{\linewidth}{!}{
  \begin{tabular}
  {p{0.8in}|p{6.2in}} \toprule
 \textbf{Tasks}
		&  \textbf{Template}  \\
		\midrule
		Connectivity/\newline Cycle/ \newline Hamilton Path
		& \texttt{In an undirected graph, (i,j) means that node i and node j are connected with an undirected edge. The nodes are numbered from [P] to [P], and the edges are: \newline         ([P], [P]) , ([P], [P])...} 
	\\ \cline{1-2}
Topological Sort & \texttt{In a directed graph with [P] nodes numbered from [P] to [P]:\newline node [P] should be visited before node [P]\newline node [P] should be visited before node [P]\newline ...} 
 \\ \cline{1-2} 
Shortest Path & \texttt{In an undirected graph, the nodes are numbered from [P] to [P], and the edges are:\newline an edge between node [P] and node [P] with weight [P],\newline an edge between node [P] and node [P] with weight [P], \newline ...} 
 \\ \cline{1-2} 
Maximum Flow & \texttt{In a directed graph, the nodes are numbered from [P] to [P], and the edges are: \newline
an edge from node [P] to node [P] with capacity [P], \newline
an edge from node [P] to node [P] with capacity [P], \newline
...}
\\ \cline{1-2}
Bipartite Graph Matching & \texttt{There are [P] hosts numbered from [P] to [P], and [P] tasks numbered from [P] to [P]. Each host has a set of tasks that it is interested in: \newline Host [P] is interested in task [P].\newline Host [P] is interested in task [P].\newline ...} 
 \\  \hline
 Link Predict/ \newline Node Classify
		& \texttt{In an undirected graph, (i,j) means that node i and node j are connected with an undirected edge. The nodes are numbered from [P] to [P], and the edges are: \newline         ([P], [P]) , ([P], [P])...\newline The node attributes are: \newline Node [P], Attribute [P]\newline Node [P], Attribute [P] \newline ...}\\
\hline
  \end{tabular}}
    \caption{Prompt templates of $T_{set}$, where [P]s are placeholders that will be substituted for specific graph topology.}
    \label{tab: prompt1}
  \end{table*}

\begin{table*}[htbp]
  \footnotesize 
  \centering 
	\renewcommand{\arraystretch}{1.1}
\resizebox{\linewidth}{!}{
  \begin{tabular}
  {p{0.8in}|p{6.2in}} \toprule
 \textbf{Tasks}
		&  \textbf{Template}  \\
		\midrule
		Connectivity/\newline Cycle/ \newline Hamilton Path
		& \texttt{In an undirected graph, (i,j) means that node i and node j are connected with an undirected edge. The nodes are numbered from [P] to [P], and the edges are presented in an adjacent list format: \newline         [P] <-> [P], [P], [P],...
        \newline
        [P] <-> [P], [P], [P],...\newline...}
	\\ \cline{1-2}
Topological Sort & \texttt{In a directed graph with [P] nodes numbered from [P] to [P]:\newline node [P] should be visited before node [P], [P], [P],...\newline node [P] should be visited before node [P], [P], [P],...\newline ...} 
 \\ \cline{1-2} 
Shortest Path & \texttt{In an undirected graph, the nodes are numbered from [P] to [P], and the edges are presented in an adjacent list format:\newline node [P] is connected to: node [P] with distance: [P], node [P] with distance: [P],... \newline node [P] is connected to: node [P] with distance: [P], node [P] with distance: [P],... \newline ...} 
 \\ \cline{1-2} 
Maximum Flow & \texttt{In a directed graph, the nodes are numbered from [P] to [P], and the edges are presented in an adjacent list format:\newline node [P] is connected to: node [P] with capacity: [P], node [P] with capacity: [P],... \newline node [P] is connected to: node [P] with capacity: [P], node [P] with capacity: [P],... \newline ...}
\\ \cline{1-2}
Bipartite Graph Matching & \texttt{There are [P] hosts numbered from [P] to [P], and [P] tasks numbered from [P] to [P]. Each host has a set of tasks that it is interested in: \newline Host [P] is interested in tasks [P], [P], [P],....\newline Host [P] is interested in task [P], [P], [P],....\newline ...} 
 \\  \hline
 Link Predict/ \newline Node Classify
		& \texttt{In an undirected graph, (i,j) means that node i and node j are connected with an undirected edge. The nodes are numbered from [P] to [P], and the edges are presented in an adjacent list format: \newline         [P] <-> [P], [P], [P],...
        \newline
        [P] <-> [P], [P], [P],...\newline... \newline The node attributes are: \newline Node [P], Attribute [P]\newline Node [P], Attribute [P] \newline ...}\\
\hline
  \end{tabular}}
    \caption{Prompt templates of $T_{list}$, where [P]s are placeholders that will be substituted for specific graph topology.}
    \label{tab: prompt2}
  \end{table*}
\begin{table*}[htbp]
  \footnotesize 
  \centering 
	\renewcommand{\arraystretch}{1.1}
\resizebox{\linewidth}{!}{
  \begin{tabular}
  {p{0.8in}|p{6.2in}} \toprule
 \textbf{Tasks}
		&  \textbf{Template}  \\
		\midrule
		Connectivity/\newline Cycle/ \newline Hamilton Path
		& \texttt{In an undirected graph, (i,j) means that node i and node j are connected with an undirected edge. The nodes are numbered from [P] to [P], and the edges are represented in an adjacent matrix format \newline            :\ \  \ node0\ \ \ \ 1\ \ \ \ 2...
        \newline node0\ \ [P]\ \ [P]\ \ [P],...\newline node1\ \ [P]\ \ [P]\ \ [P],...\newline
        node2\ \ [P]\ \ [P]\ \ [P],...\newline
        ...}
	\\ \cline{1-2}
Topological Sort & \texttt{In a directed graph with [P] nodes numbered from [P] to [P], the edges are represented in an adjacent matrix format \newline            :\ \  \ node0\ \ \ \ 1\ \ \ \ 2...
        \newline node0\ \ [P]\ \ [P]\ \ [P],...\newline node1\ \ [P]\ \ [P]\ \ [P],...\newline
        node2\ \ [P]\ \ [P]\ \ [P],...\newline
        ...} 
 \\ \cline{1-2} 
Shortest Path/\newline Maximum Flow & \texttt{In an undirected graph, the nodes are numbered from [P] to [P], and the edges are represented in an adjacent matrix format with weights \newline            :\ \  \ node0\ \ \ \ 1\ \ \ \ 2...
        \newline node0\ \ [P]\ \ [P]\ \ [P],...\newline node1\ \ [P]\ \ [P]\ \ [P],...\newline
        node2\ \ [P]\ \ [P]\ \ [P],...\newline
        ...} 
 \\ \cline{1-2} 
Bipartite Graph Matching & \texttt{There are [P] hosts numbered from [P] to [P], and [P] tasks numbered from [P] to [P]. Each host has a set of tasks that it is interested in \newline            :\ \  \ Task0\ \ \ \ 1\ \ \ \ 2...
        \newline Host0\ \ [P]\ \ [P]\ \ [P],...\newline Host1\ \ [P]\ \ [P]\ \ [P],...\newline
        Host2\ \ [P]\ \ [P]\ \ [P],...\newline
        ...} 
 \\  \hline
 Link Predict/ \newline Node Classify
		& \texttt{In an undirected graph, (i,j) means that node i and node j are connected with an undirected edge. The nodes are numbered from [P] to [P], and the edges are represented in an adjacent matrix format \newline            :\ \  \ node0\ \ \ \ 1\ \ \ \ 2...
        \newline node0\ \ [P]\ \ [P]\ \ [P],...\newline node1\ \ [P]\ \ [P]\ \ [P],...\newline
        node2\ \ [P]\ \ [P]\ \ [P],...\newline
        ...\newline The node attributes are: \newline Node [P], Attribute [P]\newline Node [P], Attribute [P] \newline ...}\\
\hline
  \end{tabular}}
    \caption{Prompt templates of $T_{mat}$, where [P]s are placeholders that will be substituted for specific graph topology.}
    \label{tab: prompt3}
  \end{table*}
\clearpage

\end{document}